\documentclass[10pt,twocolumn,letterpaper]{article}

\usepackage{cvpr}
\usepackage{times}
\usepackage{epsfig}
\usepackage{graphicx}
\usepackage{amsmath}
\usepackage{amssymb}
\usepackage{subcaption}
\usepackage{url}            
\usepackage{booktabs}       
\usepackage{amsfonts}       
\usepackage{nicefrac}       
\usepackage{microtype}      
\usepackage{lipsum}
\usepackage{graphicx}
\usepackage{caption}
\usepackage{subcaption}
\usepackage{xcolor}
\usepackage{amssymb}
\usepackage{amsfonts}
\usepackage{amsmath}
\usepackage{makecell}
\usepackage{float}
\usepackage{blindtext}
\usepackage{wrapfig}
\usepackage{enumitem}
\usepackage{multirow}

\usepackage[pagebackref=true,breaklinks=true,letterpaper=true,colorlinks,bookmarks=false]{hyperref}

\newcommand{\colornote}[3]{{\color{#1}\bf{#2: #3}\normalfont}}

\newcommand {\yi}[1]{\colornote{blue}{Yi}{#1}}

\newcommand {\mianlun}[1]{\colornote{red}{Mianlun}{#1}}

\cvprfinalcopy 

\begin{document}

\title{A Deep Emulator for Secondary Motion of 3D Characters}

\let\oldtwocolumn\twocolumn
\renewcommand\twocolumn[1][]{%
    \oldtwocolumn[{#1}{
    \begin{center}
           \includegraphics[width=0.9\textwidth]{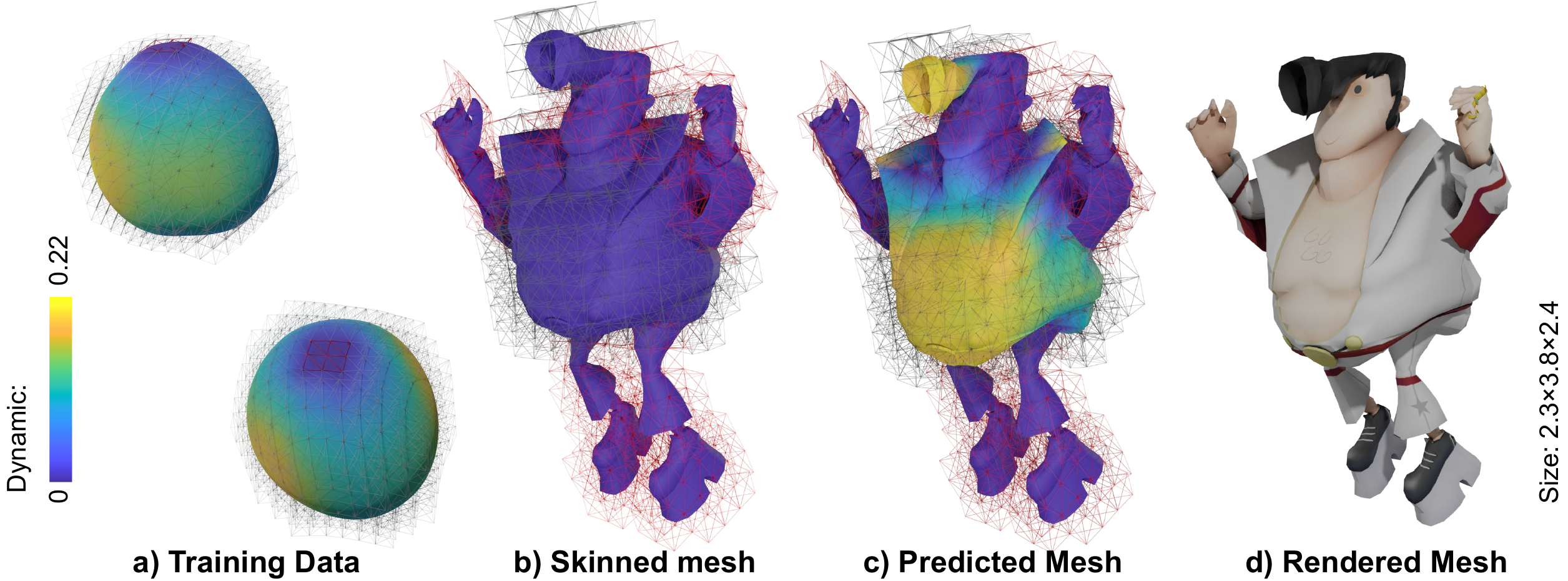}
           \captionof{figure}{a) Our method is trained on a primitive (sphere) dataset with deformation dynamics but generalizes across topology varying 3D characters. The uniform volumetric mesh surrounding the surface mesh is used for prediction, where the red vertices are set to be constraints. b) At inference time, the input is an artist-specified skinned mesh without dynamics. c) Our neural network predicts the dynamic mesh with secondary motion. d) The surface mesh is rendered with textures. 
 }
           \label{teaser}
        \end{center}
    }]
}

\author{Mianlun Zheng$^1$, Yi Zhou$^2$, Duygu Ceylan$^2$, Jernej Barbi\v{c}$^1$\\
$^1$University of Southern California, $^2$Adobe Research\\
{\tt\small \{mianlunz, jnb\}@usc.edu, \{yizho, ceylan\}@adobe.com}
}

\maketitle

\begin{abstract}

   Fast and light-weight methods for animating 3D characters are desirable in various applications such as computer games. We present a learning-based approach to enhance skinning-based animations of 3D characters with vivid secondary motion effects. We design a neural network that encodes each local patch of a character simulation mesh where the edges implicitly encode the internal forces between the neighboring vertices. The network emulates the ordinary differential equations of the character dynamics, predicting new vertex positions from the current accelerations, velocities and positions. Being a local method, our network is independent of the mesh topology and generalizes to arbitrarily shaped 3D character meshes at test time. We further represent per-vertex constraints and material properties such as stiffness, enabling us to easily adjust the dynamics in different parts of the mesh. We evaluate our method on various character meshes and complex motion sequences. Our method can be over 30 times more efficient than ground-truth physically based simulation, and outperforms alternative solutions that provide fast approximations.\footnote{Video results are available at \url{https://zhengmianlun.github.io/publications/deepEmulator.html}.}
\end{abstract}


\section{Introduction}

Fast and light-weight methods for animating 3D characters are desirable in various applications including computer games and film visual effects. Traditional skinning-based mesh deformation provides a fast geometric approach but often lacks realistic dynamics. On the other hand, physically-based simulation can add plausible secondary motion to skinned animations, augmenting them with visually realistic and vivid effects, but at the cost of heavy computation. 

Recent research has explored deep learning methods 
to approximate physically-based simulation in a much 
more time-efficient manner. While some approaches have focused on accelerating specific parts
 of the simulation~\cite{luo2018nnwarp, fulton2019latent, meister2020deep}, others have 
proposed end-to-end solutions that predict dynamics directly from mesh based features
~\cite{bailey2018fast, holden2019subspace,holden2019subspace,santesteban2020softsmpl}. While 
demonstrating impressive results, these methods still have some limitations. Most of them 
assume a fixed mesh topology and thus need to train different networks for different character
meshes. Moreover, in order to avoid the computational complexity of training networks on high
 resolution meshes, some methods operate on reduced subspaces with limited degrees of freedom, leading to low accuracy.

In this paper, we propose a deep learning approach to predict secondary motion, i.e., the deformable dynamics of given skinned animations of 3D characters. Our method addresses the shortcomings of the recent learning-based approaches by designing a network architecture that can reflect the actual underlying physical process.
Specifically, our network models the simulation using a volumetric mesh consisting of uniform tetrahedra surrounding the character mesh, where the mesh edges encode the internal forces that depend on the current state (i.e., displacements, velocities, accelerations), material properties (e.g., stiffness), and constraints on the vertices. Mesh vertices  encode the inertia. 
Motivated by the observation that within a short time instance the secondary dynamics of a vertex is mostly affected by its current state, as well as the internal forces due to its neighbors, our network operates on local patches of the volumetric mesh. In addition to avoiding the computational complexity of encoding high resolution character meshes as large graphs, this also enables our method to be applied to any character mesh, independent of its topology. Finally, our network encodes per-vertex material properties and constraints, giving the user the ability to easily prescribe varying properties to different parts of the mesh to control the dynamic behaviour.

As a unique benefit of the generalization capability of our model, we demonstrate that it is not necessary to construct a massive training dataset of complex meshes and motions. Instead, we construct our training data from primitive geometries, such as a volumetric mesh of a sphere. Our network trained on this dataset can generate detailed and visually plausible secondary motions on much more complex 3D characters during testing. 
By assigning randomized motions to the primitives during training, we are able to let the local patches cover a broad motion space, which improves the network's online predictions in unseen scenarios.

We evaluate our method on various character meshes and complex motion sequences. We demonstrate visually plausible and stable secondary motion while being over 30 times faster than the implicit Euler method commonly used in physically-based simulation. We also provide comparisons to faster methods such as the explicit central differences method and other learning-based approaches that utilize graph convolutional networks. Our method outperforms those approaches both in terms of accuracy and robustness.


\section{Related Work}

\subsection{Physically based simulation methods}
Complementing skinning-based animations with secondary motion is a well-studied problem. Traditional approaches resort to using physically-based simulation~\cite{Zhang:CompDynamics:2020, Wang:2020:ACS}. However, it is well-known that physically based methods often suffer from computational complexity. Therefore, in the last decade, a series of methods were proposed to accelerate the computation process, including example-based dynamic skinning~\cite{shi2008example}, efficient elasticity calculation~\cite{mcadams2011efficient}, formulation of motion equations in the rig subspace~\cite{hahn2012rig,hahn2013efficient}, and the coupling of the skeleton dynamics and the soft body dynamics~\cite{liu2013simulation}. These approaches still have some limitations such as robustness issues due to explicit integration, or unnatural deformation effects due to remeshing, while our method is much more robust in handling various characters and complex motions.


\subsection{Learning based methods}

Grzeszczuk et al.~\cite{grzeszczuk1998neuroanimator} presented one of the earliest works that demonstrated the possibility of replacing numerical computations with a neural network. Since then research in this area has advanced, especially in the last few years. While some approaches have presented hybrid solutions where a neural network replaces a particular component of the physically based simulation process, others have presented end-to-end solutions.

In the context of hybrid approaches, plug-in deep neural networks were applied in combination with the Finite Elements Method (FEM), to help accelerate the simulation. 
For example, the node-wise NNWarp~\cite{luo2018nnwarp} was proposed to efficiently map the linear nodal displacements to nonlinear ones. 
Fulton et al.\cite{fulton2019latent} utilized an autoencoder to project the target mesh to a lower dimensional space to increase the computation speed. 
Similarly, Tan et al.~\cite{tan2020realtime} designed a CNN-based network for dimension reduction to accelerate thin-shell deformable simulations. 
Romero et al.~\cite{ROCP20} built a data-driven statistical model to kinematically drive the FEM mechanical simulation. Meister et al. ~\cite{meister2020deep} explored the use of neural networks to accelerate the time integration step of the Total Lagrangian Explicit Dynamics (TLED) for complex soft tissue deformation simulation. Finally, Deng et al.~\cite{deng2020alternating} modeled the force propagation mechanism in their neural networks.
Those approaches improved efficiency but at the cost of accuracy and are not friendly to end users who are not familiar with physical techniques. Ours, instead, allows the user to adjust the animation by simply painting the constraints and stiffness properties. 

End-to-end approaches assume the target mesh is provided as input and directly predict the 
dynamics behaviour. For instance, Bailey et al.~\cite{bailey2018fast} enriched the real-time 
skinning animation by adding the nonlinear deformations learned from film-quality character 
rigs. The work of Holden et al~\cite{holden2019subspace} first trained an autoencoder to 
reduce the simulation space and then learned to efficiently approximate the dynamics projected 
to the subspace. Similarly,  SoftSMPL~\cite{santesteban2020softsmpl} modeled the realistic 
soft-tissue dynamics based on a novel motion descriptor and a neural-network-based recurrent 
regressor that ran in the nonlinear deformation subspace extracted from an autoencoder. While all these approaches presented impressive results, their main drawback was the assumption of a fixed mesh topology requiring different networks to be trained for different meshes. Our approach, on the other hand, operates at a local patch level and can therefore generalize to different meshes at test time. 

Lately, researchers started to utilize the Graph Convolutional Network (GCN) for simulation tasks due to its advantage in handling topology-free graphs. The GCN encodes the vertex positional information and aggregates the latent features to a certain node by using the propagation rule. For particle-based systems, graphs are constructed based on the local adjacency of the particles at each frame and fed into GCNs ~\cite{li2018learning,ummenhofer2019lagrangian, sanchez2020learning,de2020combining}. Concurrently, Pfaff et al.~\cite{pfaff2020learning} proposed a GCN for surface mesh-based simulation.  
While these GCN models interpret the mesh dynamics prediction as a general spatio-temporal problem, we incorporate physics into the design of our network architecture, e.g. inferring latent embedding for inertia and internal forces, which enables us to achieve more stable and accurate results (Section~\ref{comparison}).


\section{Method}
\begin{figure}[!t]
    \centering
    \includegraphics[width=1.0\linewidth]{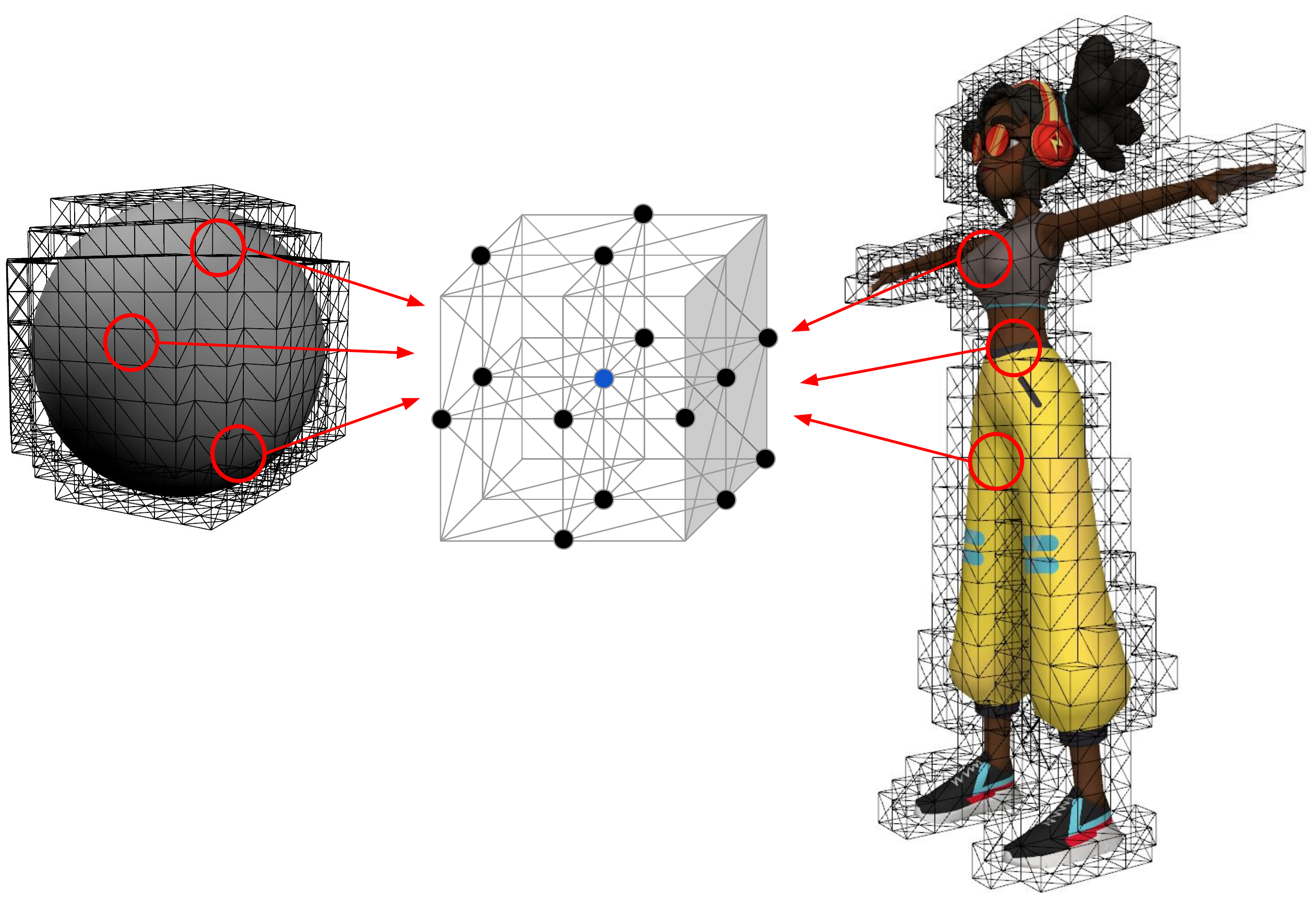}
    \caption{The tetrahedral simulation mesh and the embedded surface mesh. The local patch consists of a center vertex and its neighbors, defined as the vertices of the tetrahedra touching the center vertex.}
    \label{tet_mesh}
\end{figure}

Given a 3D character and its primary motion sequence obtained using standard linear blend skinning techniques~\cite{skinningcourse:2014}, we first construct a volumetric (tetrahedral) mesh and a set of barycentric weights 
to linearly embed the vertices of the character's surface mesh into the volumetric mesh~\cite{James:2004:Squashing}, as shown in Figure~\ref{tet_mesh}.
Our network operates on the volumetric mesh and predicts the updated vertex positions with deformable dynamics (also called the secondary motion) at each frame given the primary motion, the constraints and the material properties. The updated volumetric mesh vertex positions then drive the original surface mesh via the barycentric embedding, and the surface mesh is used for rendering; such a setup is very common and standard in computer animation.

We denote the reference tetrahedral mesh and its number of vertices by $X$ and $n,$ respectively. The skinned animation (primary motion) is represented as a set of time-varying positions $\mathbf{x}\in \mathbb{R}^{3n}$. Similarly, we denote the predicted dynamic mesh by $U$ and its positions by ${\mathbf{u}}\in \mathbb{R}^{3n}$.

Our method additionally encodes mass ${\mathbf{m}}\in \mathbb{R}^{n}$ and stiffness ${\mathbf{k}}\in \mathbb{R}^{n}$ properties. The stiffness is represented as Young's modulus. By painting different material properties per vertex over the mesh, users can control the dynamic effects, namely the deformation magnitude. 

In contrast to previous works~\cite{santesteban2020softsmpl,pfaff2020learning} which trained neural networks directly on the surface mesh, we choose to operate on the volumetric mesh for several reasons. First, volumetric meshes provide a more efficient coarse representation and can handle character meshes that consist of multiple disconnected components. For example, in our experiments the ``Michelle'' character (see Figure~\ref{tet_mesh}) consists of $14k$ vertices whereas the corresponding volumetric mesh only has $1k$ vertices. In addition, the ``Big Vegas'' character mesh (see Figure~\ref{teaser}) has eight disconnected components, requiring the artist to build a watertight mesh first if using a method that learns directly on the surface mesh. Furthermore, volumetric meshes not only capture the surface of the character but also the interior, leading to more accurate learning of the internal forces. Finally, we use a uniformly voxelized mesh subdivided into tetrahedra as our volumetric mesh, which enables our method to generalize across character meshes with varying shapes and resolutions.

Next, we will first explain the motion equations in physically-based simulation and then discuss our method in detail, drawing inspiration from the physical process.

\subsection{Physically-based Motion Equations}

In constraint-based physically-based simulation~\cite{baraff2001physically}, the equations of motion are
\begin{gather}
\label{eqn:motion}
\mathbf{M}\ddot{\mathbf{u}}+\mathbf{D} \dot{\mathbf{u}}+ \mathbf{f}_{int}(\mathbf{u}) = \mathbf{0} \\
\text{subject to } \mathbf{C}\mathbf{u} = \mathbf{Sx}(t) \nonumber,
\end{gather}
where $\mathbf{M}\in \mathbb{R}^{3n\times 3n}$  is the diagonal (lumped) mass matrix (as commonly employed in interactive applications),
$\mathbf{D}$ is the Rayleigh damping matrix, and ${\mathbf{u}\in \mathbb{R}^{3n}}$, $\dot{\mathbf{u}}\in \mathbb{R}^{3n}$  and $\ddot{\mathbf{u}}\in \mathbb{R}^{3n}$ represent the positions, velocities and accelerations, respectively. The quantity $\mathbf{f}_{int}(\mathbf{u})$ represents the internal elastic forces. 
Secondary dynamics occurs because the constraint part of the mesh ``drives" the free part of the mesh.
Constraints are specified via the constraint matrix $\mathbf{C}$ and the selection matrix $\mathbf{S}$. In order to leave room for secondary dynamics for 3D characters, we typically do not constrain \emph{all} the vertices of the mesh, but only a subset. For example, in the Big Vegas example (see Figure~\ref{teaser}), we constrain the legs, the arms and the core inside the torso and head, but do not constrain the belly and hair, so that we can generate secondary dynamics in those unconstrained regions.

One approach to timestep Equation~\ref{eqn:motion} is to use an explicit integrator, such as central differences: 
\begin{align}\label{eqn:explicit_integration}
\begin{split}
\dot{\mathbf{u}}(t+1) &= \dot{\mathbf{u}}(t) +\frac{\ddot{\mathbf{u}}(t)+\ddot{\mathbf{u}}(t+1)}{2}\Delta t,\\
\mathbf{u}(t+1) &= \mathbf{u}(t) + \dot{\mathbf{u}}(t)\Delta t + \ddot{\mathbf{u}}(t)\frac{\Delta t^2}{2},\\
\end{split}
\end{align}
where $t$ and $t+1$ denote the state of the mesh in the current and next frames, respectively, and $\Delta t$ is the timestep. While the explicit integration is fast, it suffers from stability issues. Hence, the slower but stable implicit backward Euler integrator is often preferred in physically-based simulation~\cite{Baraff:1998:LSI}:
\begin{align}\label{eqn:implicit_integration}
\begin{split}
\dot{\mathbf{u}}(t+1) &= \dot{\mathbf{u}}(t) + \ddot{\mathbf{u}}(t+1)\Delta t,\\
\mathbf{u}(t+1) &= \mathbf{u}(t) + \dot{\mathbf{u}}(t+1)\Delta t.\\
\end{split}
\end{align}

We propose to approximate implicit integration as
\begin{align}
\label{eqn:deep_integration}
\begin{split}
\dot{\mathbf{u}}(t+1) &= \dot{\mathbf{u}}(t) + \\ &f_{\theta}\Bigl(\mathbf{u}(t), \dot{\mathbf{u}}(t), \ddot{\mathbf{u}}(t),
\mathbf{x}(t),
\dot{\mathbf{x}}(t),
\ddot{\mathbf{x}}(t), \mathbf{m}, \mathbf{k}\Bigr)\Delta t,\\
\mathbf{u}(t+1) &= \mathbf{u}(t) + \dot{\mathbf{u}}(t+1)\Delta t,
\end{split}
\end{align}
where $f$ is a differentiable function constructed as a neural network with learned parameters $\theta$.

\subsection{Network design}
\begin{figure}[!t]
    \centering
    \includegraphics[width=0.8\linewidth]{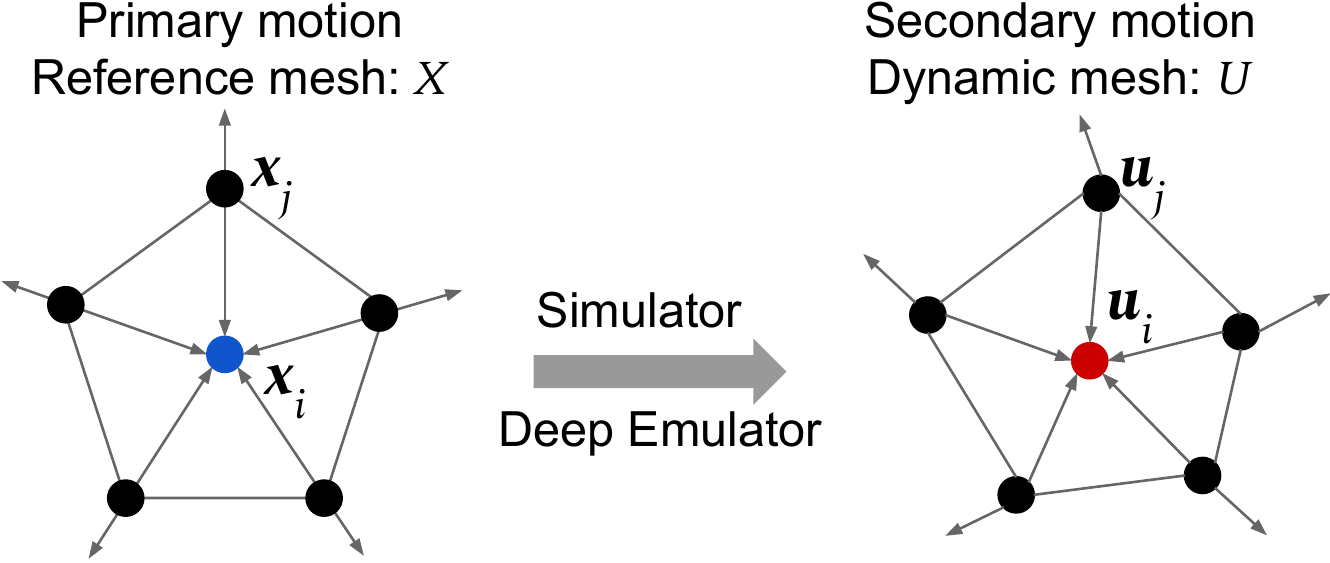}
    \caption{The input reference mesh $X$ and the target dynamic mesh $U$. We draw the meshes in 2D for convenience. }
    \label{mesh}
\end{figure}

As shown in Equation~\ref{eqn:motion}, predicting the secondary dynamics entails solving for $3n$ degrees of freedom for a mesh with $n$ vertices. Hence,  directly approximating $f_{\theta}$ in Equation~\ref{eqn:deep_integration} to predict all the degrees of freedom at once would lead to a huge and impractical network, which would furthermore not be applicable to input meshes with varying number of vertices and topologies. Inspired by the intuition that within a very short time moment, the motion of a vertex is mostly affected by its own inertia and the internal forces from its neighboring vertices, we design our network to operate on a local patch instead. As illustrated in Figure~\ref{mesh}, the 1-ring local patch consists of one center vertex along with its immediate neighbors in the volumetric mesh. Even though two characters might have very different mesh topologies, as shown in Figure~\ref{tet_mesh}, their local patches will often be more similar, boosting the generalization ability of our network.

The internal forces are caused by the local stress, and the aggregation of the internal forces acts to pull the vertices to their positions in the reference motion, to reduce the elastic energy.  Thus, the knowledge of the per-edge deformation and the per-vertex reference motion are needed for secondary motion prediction.

\begin{figure}[!htbp]
    \centering
    \includegraphics[width=1.0\linewidth]{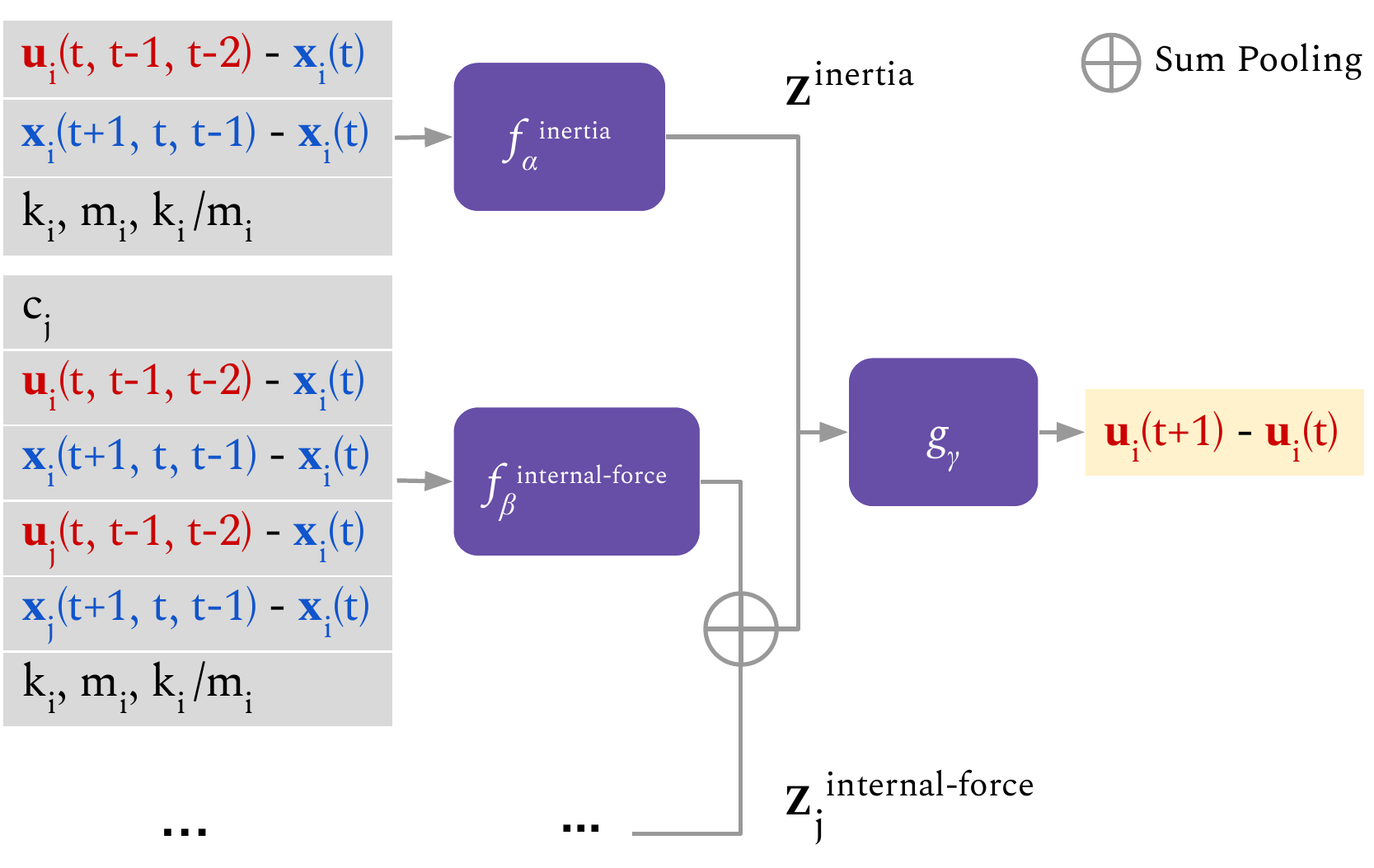}
    \caption{Our network architecture.}
    \label{network}
\end{figure}

Hence, we propose to emulate this process as follows:
\begin{align}\label{eqn:deep_solution}
\begin{split}
&\mathbf{z}_i^{\textrm{inertia}}=f_{\alpha}^{\textrm{inertia}}(\mathbf{u}_i, \mathbf{x}_i, k_i, m_i), \\
&\mathbf{z}_{i,j}^{\textrm{internal\_force}}=f_{\beta}^{\textrm{internal\_force}}(\mathbf{u}_{i,j}, \mathbf{x}_{i,j}, k_i, m_i),\\
&\ddot{\mathbf{u}}_i= g_{\gamma}\bigl(\mathbf{z}_i^{\textrm{\textrm{inertia}}}, \sum_{j\in \mathbb{N}_i}\mathbf{z}_{i,j}^{\textrm{internal\_force}}\bigr),
\end{split}
\end{align}
where $f_{\alpha}^{\textrm{inertia}}$, $f_{\beta}^{\textrm{internal\_force}}$ and $g_{\gamma}$ are three different multi-layer perceptrons (MLPs) as shown in Figure~\ref{network}, $\mathbb{N}_i$ are neighboring
vertices of $i$ (excluding $i$),
and the double indices $i,j$ denote
the central vertex $i$ and a neighbor $j.$
Quantities
$\mathbf{z}_i^{\textrm{inertia}}$ and $\mathbf{z}_i^{\textrm{internal\_force}}$ are high dimensional latent vectors that represent an embedding for inertia dynamics and the internal forces from each neighboring vertex, respectively. Perceptron $g_{\gamma}$ receives the concatenation of $\mathbf{z}_i^{\textrm{inertia}}$ and the sum of $\mathbf{z}_i^{\textrm{internal\_force}}$ to predict the final acceleration of a vertex $i$. 
 In practice, for simplicity, we train $g_{\gamma}$ to directly predict $\dot{\mathbf{u}}(t+1)\Delta t = \mathbf{u}(t+1)-\mathbf{u}(t)$ since  we assume a fixed timestep of $\Delta t$ in our experiments. 

We implement all the three MLPs with four hidden fully connected layers activated by the Tanh function, and one output layer. During training, we provide the ground truth positions in the dynamic mesh as input. During testing, we provide the predictions of the network as input in a recurrent manner.
Next, we discuss the details of these components.

\vspace{-0.3cm}
\paragraph{MLP $f_\alpha^{\textrm{inertia}}$:}
This perceptron focuses on the center vertex itself, encoding the  ``self-inertia" information. That is, the center vertex tends to continue its current motion, driven by both the velocity and acceleration.  The input to $f_\alpha^{\textrm{inertia}}$ is the position of the center vertex in the last three frames both on the dynamic and skinned mesh, $\mathbf{u}(t)$,$\mathbf{u}(t-1)$,$\mathbf{u}(t-2)$ and $\mathbf{x}(t+1)$,$\mathbf{x}(t)$, $\mathbf{x}(t-1)$, as well as its material properties, $k_i, m_i, k_i/m_i$. The positions are represented in local coordinates with respect to $\mathbf{x}(t)$, the current position of the center vertex in the reference motion. The positions in the last three frames implicitly encode the velocity and the acceleration. Since we know that the net force applied on the central vertex is divided by its mass in Equation~\ref{eqn:deep_integration} and it is relatively hard for the network to learn multiplication or division, we also include $k_i/m_i$ explicitly in the input. The hidden layer and output size is 64.
 
\vspace{-0.3cm}
\paragraph{MLP $f_\beta^{\textrm{internal\_force}}$:} For an unconstrained center vertex $i$, perceptron $f_\beta^{\textrm{internal\_force}}$ encodes the ``internal forces'' contributed by its neighbors. The input to the MLP is similar to $f_\alpha^{\textrm{inertia}},$ except that we provide information both for the center vertex as well as its neighbors. For each neighboring vertex $j$, we also provide the constraint information $c_j$ ($c_j = 0$ if a free vertex; $c_j = 1$ if constrained).   
Each $f_\beta^{\textrm{internal\_force}}$ provides a latent vector for the central vertex. 
The hidden layer and output size is 128.

\vspace{-0.3cm}
\paragraph{MLP $g_\gamma$:}
This module receives the concatenated outputs from $f_\alpha^{\textrm{inertia}}$ and the aggregation of $f_\beta^{\textrm{internal\_force}}$, and predicts the final displacement of the central vertex $i$ in the dynamic mesh. The input and hidden layer size is 192. 

We train the final network with the mean square error loss:
\begin{align}\label{eqn:loss}
\begin{split}
l = \frac{1}{n}\sum_{i}^n ||\dot{\mathbf{u}}_{i}(t+1) - \dot{\mathbf{u}}'_{i}(t+1)||_2^2,
\end{split}
\end{align}
where $\dot{\mathbf{u}}'_{i}(t+1)$ is the ground truth. We adopted the Adam optimizer for training, with a learning rate starting from 0.0001 along with a decay factor of 0.96 at each epoch.  

\subsection{Training Primitives}
Because our method operates on local patches, it is not necessary to train it on complex character meshes. In fact, we found that a training dataset constructed by simulating basic primitives, such as a sphere (under various motions and material properties), is sufficient to generalize to various character meshes at test time. Specifically, we generate random motion sequences by prescribing random rigid body motion of a constrained beam-shaped core inside the spherical mesh. The motion of this rigid core excites dynamic deformations in the rest of the sphere volumetric mesh. Each motion sequence starts by applying, to the rigid core, a random acceleration and angular velocity with respect to a random rotation axis. Next, we reverse the acceleration so that the primitive returns back to its starting position, and let the primitive's secondary dynamics oscillate out for a few frames. While the still motions ensure that we cover the cases where local patches are stationary (but there is still residual secondary dynamics from primary motion), the random accelerations help to sample a diverse set of motions of local patches as much as possible. Doing so enhances the networks's prediction stability.
%


\section{Experiments}
 In this section, we show qualitative and quantitative results of our method, as well as comparisons to other methods. We also run an ablation study to verify why explicitly providing the position information on the reference mesh as input is necessary.
 
\subsection{Dataset and evaluation metrics}\label{train}
For training, we use a uniform tetrahedral mesh of a sphere. We generate $80$ random motion sequences at 24 fps, using the Vega FEM simulator~\cite{Vega, sin2013vega}. For each motion sequence, we use seven different material settings. Each motion sequence consists of 456 frames resulting in a total of 255k frames in our training set.

We evaluate our method on 3D character animations obtained from Adobe's Mixamo dataset~\cite{mixamo}. Neither the character meshes nor the primary motion sequences are seen in our training data. We create test cases for five different character meshes as listed in Table~\ref{tab:time_result} and 15 motions in total. The volumetric meshes for the test characters use the same uniform tetrahedron size as our training data. For all the experiments, we report three types of metrics:

\begin{itemize}[leftmargin=0.4cm]
    \vspace{-0.2cm}
    \item Single-frame RMSE: We measure the average root-mean-square error (RMSE) between the prediction and the ground truth over all frames, while providing the ground truth positions  of the previous frames as input.
    \vspace{-0.2cm}
    \item Rollout RMSE: We provide the previous predictions of the network as input to the current frame in a recurrent manner and measure the average RMSE between the prediction and the ground truth over all frames.
    \vspace{-0.2cm}
    \item $E_{elastic}[min, stdev, max]$: We use the concept of elastic energy in physically-based simulation to detect abnormalities in the deformation sequence, or any possible mesh explosions. For each frame, we calculate the elastic energy based on the current mesh displacements with respect to its reference state.
    We list the the $min$, $max$ as well as the standard deviation ($stdev)$ to show the energy distribution across the animation. 
\end{itemize}

\subsection{Analysis of Our Method}\label{method_analysis}

\paragraph{Performance:} In Table~\ref{tab:time_result}, we show the speed $t_{ours}$ of our method, as well as that of the ground truth method $t_{GT}$ and a baseline method $t_{BL}$. For each method, we record the time to calculate the dynamic mesh but exclude other components such as initialization, rendering and mesh interpolation. 

We adopted the implicit backward Euler approach (Equation~\ref{eqn:implicit_integration}) as ground truth and the faster explicit central differences integration (Equation~\ref{eqn:explicit_integration}) as the baseline. Both our baseline and ground truth were
optimized using the deformable object simulation library, Vega FEM~\cite{Vega,sin2013vega}, and accelerated using multi-cores via Intel Thread Building Blocks (TBB), with 8 cores for assembling the internal forces and 16 cores for solving the linear system. The experiment platform is with 2.90 GHz Intel Xeon(R) CPU E5-2690 (32 GB RAM) which provides for a highly competitive baseline/ground truth implementation. 
We ran our trained model on a GeForce RTX 2080 graphics card (8 GB  RAM). We also tested it on CPU, without any multi-thread acceleration. 

Moreover, we also provide performance results for the same character mesh (Big Vegas) with different voxel resolutions. To handle different resolutions of testing meshes, we resize the volumetric mesh to have the local patch similar to the training data (i.e., the shortest edge length is 0.2). 

\begin{table*}[htbp]
    \centering
    \begin{tabular}{l|c|c|c|c|c}
    \hline
     character & \makecell{\# vertices \\ (tet mesh)} & \makecell{$t_{GT}$ \\ s/frame} & \makecell{$t_{BL}$ \\ s/frame} & \makecell{$t_{ours} (GPU)$ \\ s/frame} & \makecell{$t_{ours} (CPU)$ \\ s/frame}  \\
     \hline
     Big vegas & 1468 & 0.58 & 0.056 & \textbf{0.012} & 0.017\\
     Kaya & 1417 & 0.52	& 0.052 & \textbf{0.012} & 0.015\\
     Michelle & 1105 & 0.33 & 0.032 & \textbf{0.011} & 0.015\\
     Mousey & 2303 & 0.83 & 0.084 & \textbf{0.014} & 0.020 \\
     Ortiz & 1258 & 0.51 & 0.049 & \textbf{0.012} & 0.015 \\
     Big vegas & 6987 &  2.45 & 0.32 & \textbf{0.032} & 0.14 \\
     Big vegas & 10735 & 4.03 & 0.53 & \textbf{0.046} & 0.24 \\
     Big vegas & 18851 & 8.26 & 1.06 & \textbf{0.068} & 0.42\\
     Big vegas & 39684 & 24.24 & 2.96 & \textbf{0.14} & 0.89\\
     \hline
    \end{tabular}
    \caption{The running time (s/frame) of a single step (1/24 second) for the ground truth, the baseline, and our method.}
    \label{tab:time_result}
\end{table*}

Results indicate that when ran on GPU (CPU), our method is around 30 ($\sim$20) times faster than the implicit integrator and 3 ($\sim$2) times faster than the explicit integrator, per frame. Under an increasing
number of vertices, our method has an even more competitive performance. Although the explicit method has comparable speed to our method, the simulation explodes after a few frames. In practice, explicit methods require much smaller time steps, which required additional 100 sub-steps in our experiments, to achieve stable quality. We provide a more detailed report on the speed-stability relationship of explicit integration in the supplementary material.

\begin{figure}[!htbp]
    \centering
    \includegraphics[width=0.9\linewidth]{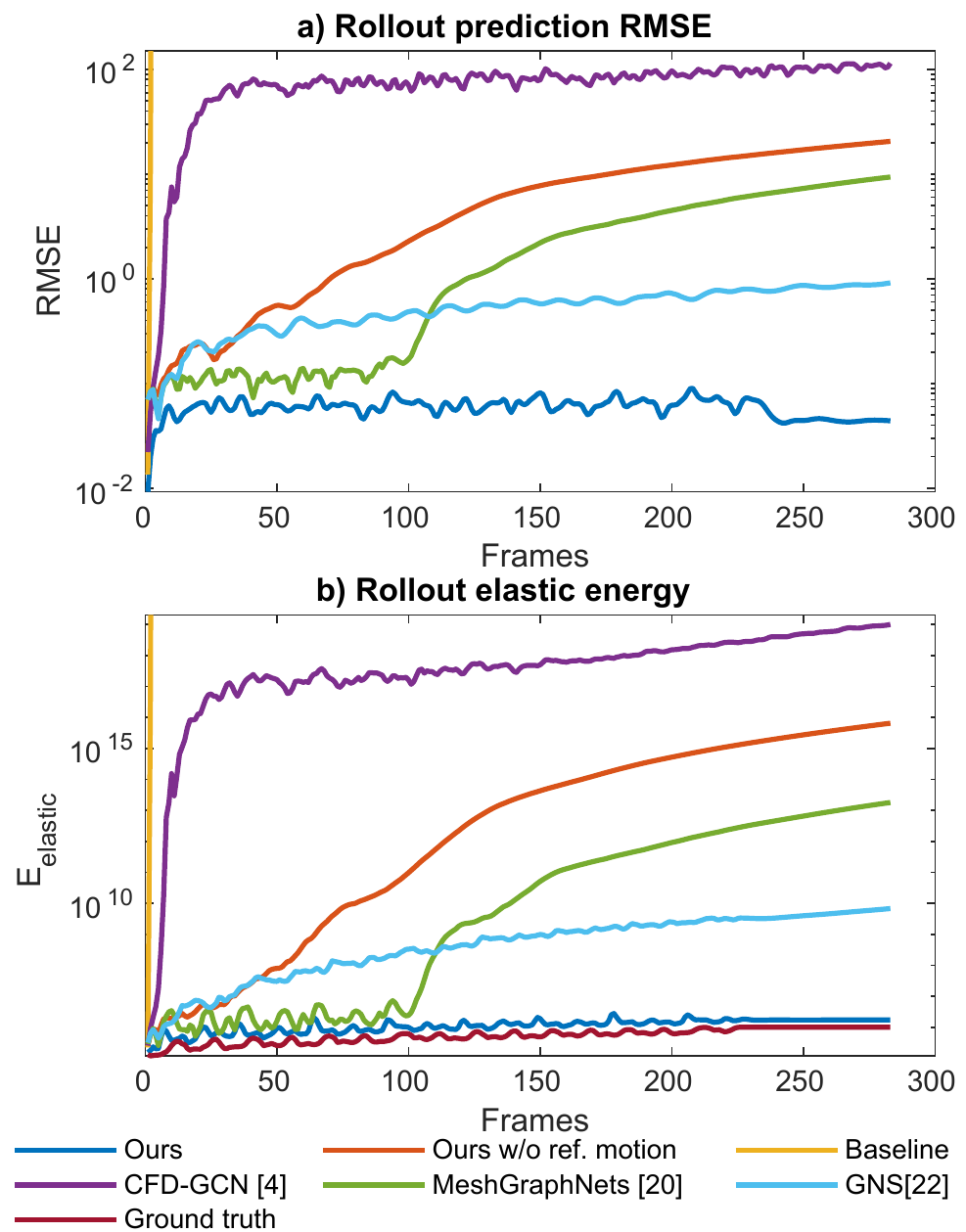}
    \caption{The rollout prediction results of our method and others, tested on the Big Vegas character with 283-frame hip hop dancing motion.}
    \label{fig:big_vegas_plot}
\end{figure}

\begin{figure*}[!htbp]
    \centering
    \includegraphics[width=1.0\linewidth]{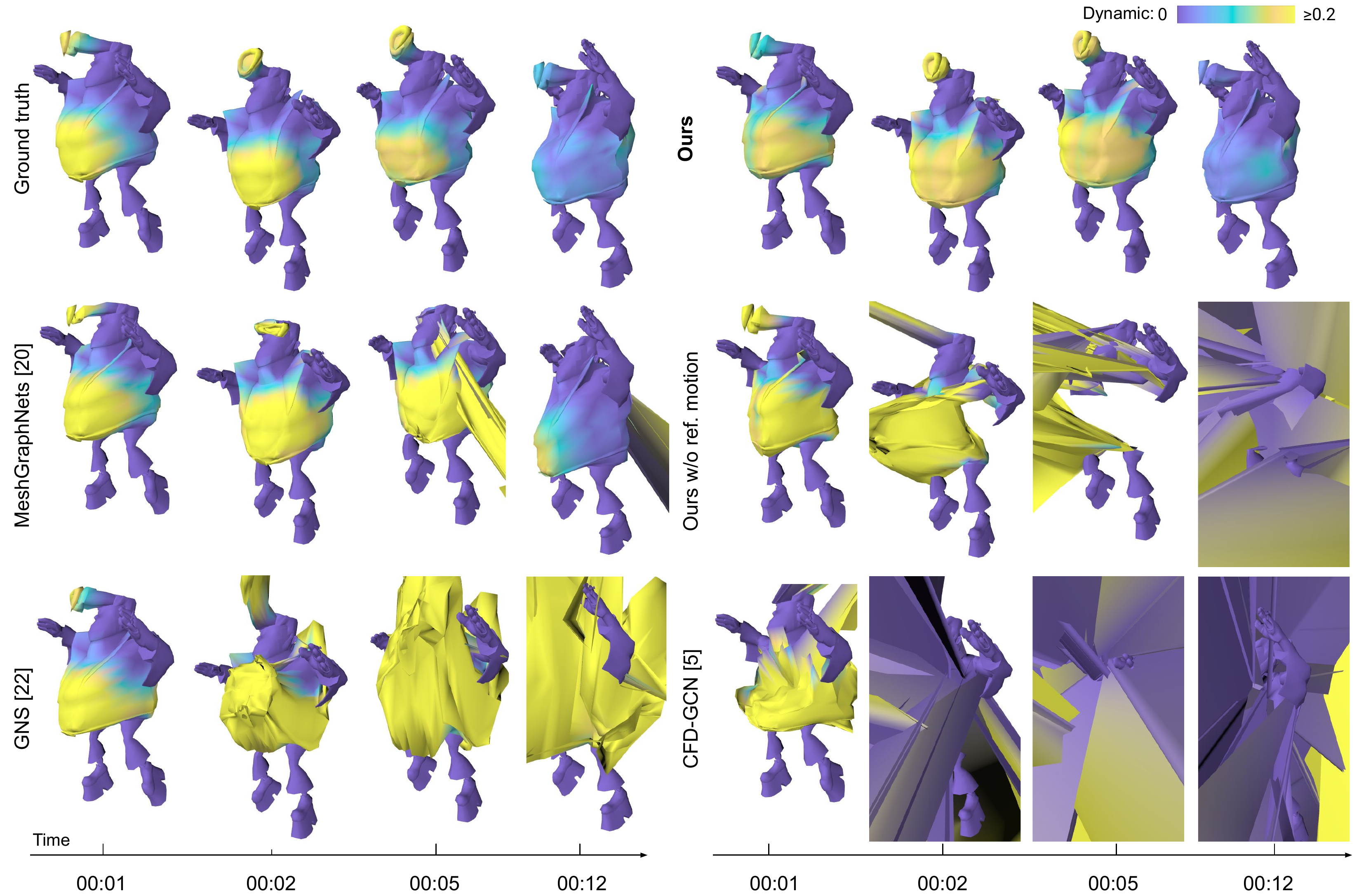}
    \caption{The rollout prediction results of our method and others tested on the Big Vegas character with 283-frame hip hop dancing motion. The baseline cannot be rendered because it explodes.}
    \label{fig:big_vegas_visual}
\end{figure*}

\vspace{-0.3cm}
\paragraph{Generalization:} We train the network on the sphere dataset and achieve a single frame RMSE of 0.0026 on the testing split of this dataset (the sphere has a radius of $2$). As listed in Table~\ref{tab:accuracy_result}, when tested on characters, our method achieves a single frame RMSE of $0.0067$, showing remarkable generalization capability (we note that the shortest edge length on the volumetric character meshes is $0.2$). The mean rollout error increases to $0.064$ after running the whole sequences due to error accumulation, but elastic energy statistics are still close to the ground truth. From the visualization of the ground truth and our results in Figure~\ref{fig:big_vegas_visual}, we can see that although the predicted secondary dynamics have slight deviation from the ground truth, they are still visually plausible. We further plot the rollout prediction RMSE and elastic energy of the Big Vegas character in Figure~\ref{fig:big_vegas_plot}. It can be seen that the prediction error remains under $0.07$, and the mean elastic energy of our method is always close to the ground truth for the whole sequence, whereas the baseline method explodes quickly. We provide such rollout prediction plots for all characters in the supplemental and the video results in supplemental material.

\begin{table*}[!htbp]
    \centering
    \begin{tabular}{l|c|c|c|c|c}
    \hline
     Methods & single frame & rollout-24 & rollout-48 & rollout-all  & \makecell{$E_{elastic}$\\$[min, stdev, max]$} \\
     \hline
     Ground truth & $\setminus $ & $\setminus $ & $\setminus $ & $\setminus$ & $[3.81E3, 4.06E5, 2.60E6]$ \\
     Our method & \textbf{0.0067} & \textbf{0.059} & \textbf{0.062} & \textbf{0.064} & $\mathbf{[4.84E3,	6.51E5, 6.32E6]}$ \\
     Ours w/o ref. motion & 0.050 & 0.20 & 0.38 & 10.09 & $[1.62E4, 6.7E16, 4.7E17]$\\
     Baseline & $\setminus $ & 7.26$E120$ & 9.63$E120$ & 17.5$E120$ & $[9.26E0, Nan, 7.22E165]$ \\
     CFD-GCN~\cite{de2020combining} & 0.040 & 41.17 & 70.55 & 110.07 & $[3.96E4, 1.1E22, 1.6E23]$ \\
     GNS~\cite{sanchez2020learning} & 0.049 & 0.22 & 0.34 & 0.54 & $[1.09E4, 2.0E11, 2.3E10]$\\
     MeshGraphNets~\cite{pfaff2020learning} & 0.050 & 0.11 & 0.43 & 4.46 & $[1.69E4, 1.1E15,1.1E14]$\\
     \hline 
    \end{tabular}
    \caption{The single-frame RMSE, rollout-24, rollout-48 and rollout-all of our method and others tested on all five characters with 15 different motions. The shortest edge length in the test meshes is 0.2. }
    \label{tab:accuracy_result}
\end{table*}

\begin{figure}[!htbp]
    \centering
    \includegraphics[width=1.0\linewidth]{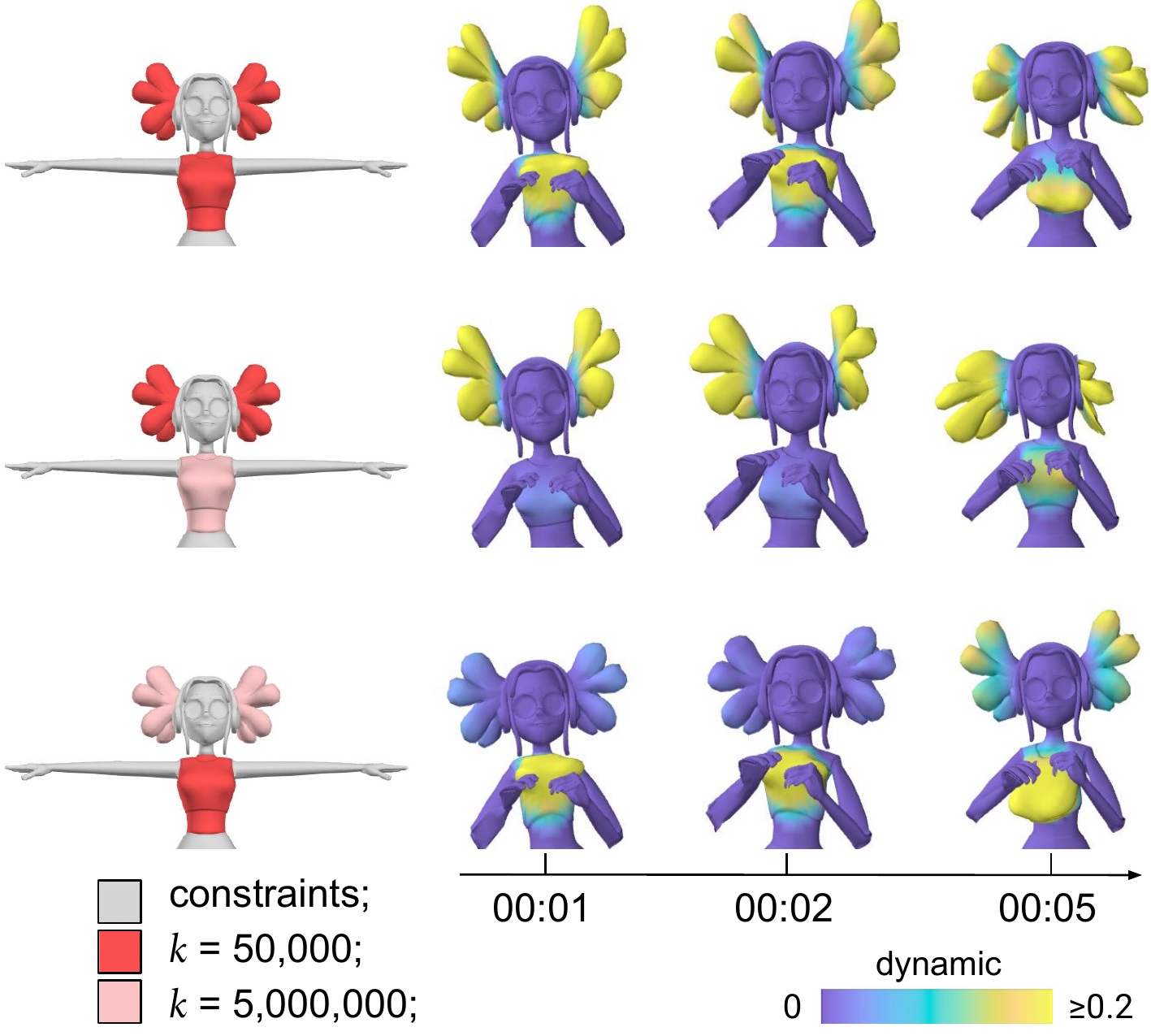}
    \caption{Non-homogeneous dynamics, tested on the Michelle character with 122-frame cross-jumps motion. We only show the upper region (see Figure~\ref{tet_mesh} for the full mesh).}
    \label{fig:michelle_inhomogeneous}
\end{figure}

\vspace{-0.3cm}
\paragraph{Non-homogeneous Dynamics:}
Figure~\ref{fig:michelle_inhomogeneous} shows how to control the dynamics by painting non-homogeneous material properties over the mesh. Varying stiffness values are painted on the hair and the breast region on the volumetric mesh. For better visualization, we render the material settings across the surface mesh in the figure. We display three different material settings, by assigning different stiffness $k$ values. Larger $k$ means stiffer material, hence the corresponding region exhibits less dynamics. In contrast, the regions with smaller $k$ show significant dynamic effects. This result demonstrates that our method correctly models the effect of material properties while providing an interface for the artist to efficiently adjust the desired dynamic effects. 

\vspace{-0.3cm}
\paragraph{Ablation study:}
\label{ablation}
To demonstrate that it is necessary to incorporate the reference mesh motion into the input features of our network, we performed an ablation study. To ensure that the constrained vertices are still driving the dynamic mesh in the absence of the reference information,  we update the positions of the constrained vertices based on the reference motion, at the beginning of each iteration. As input to our network architecture, we use the same set of features except the positions on the reference mesh. The results of ``Ours w/o ref. motion" in Table~\ref{tab:accuracy_result} and Figure~\ref{fig:big_vegas_visual} and \ref{fig:big_vegas_plot} demonstrate that this version is inferior to our original method, especially when running the network over a long time sequence. This establishes that the reference mesh is indispensable to the quality of the network's approximation.


\subsection{Comparison to Previous Work}
\label{comparison}

As discussed in Section 2, several recent particle-based physics and mesh-based deformation systems utilized graph convolutional networks (GCNs). In this section, we train these network models on the same training set as our method and test on our character meshes.
\vspace{-0.3cm}
\paragraph{CFD-GCN~\cite{de2020combining}:} We implemented our version of the CFD-GCN architecture, adopting the convolution kernel of~\cite{kipf2016semi}. However, we ignored the remeshing part because we assume that the mesh topology remains fixed when predicting secondary motion. As input, we provide the same information as our method, namely the constraint states of the vertices, the displacements and the material properties. We found that the network structure recommended in the paper resulted in a high training error. We then replaced the originally proposed ReLu activation function with the Tanh activation (as used in our method), which significantly improved the training performance. Even so, as shown in Table~\ref{tab:accuracy_result} and Figure~\ref{fig:big_vegas_plot}, the rollout prediction explodes very quickly. We speculate that although the model aggregates the features from the neighbors to a central vertex via an adjacency matrix, it treats the center and the neighboring vertices equally, whereas in reality, their roles in physically-based simulation are distinct.
\vspace{-0.3cm}
\paragraph{GNS~\cite{sanchez2020learning}:} The recently proposed GNS~\cite{sanchez2020learning} architecture is also a graph network designed for particle systems. The model first separately encodes node features and edge features in the graph and then generalizes the GraphNet blocks in~\cite{pmlr-v80-sanchez-gonzalez18a} to pass messages  across the graph. Finally, a decoder is used to extract the prediction target from the GraphNet block output. The original paper embeds the particles in a graph by adding edges between vertices under a given radius threshold. In our implementation, we instead utilized the mesh topology to construct the graph. We used two blocks in the ``processor''~\cite{sanchez2020learning} to achieve a network capacity similar to ours.
In contrast to CFD-GCN~\cite{de2020combining}, the GraphNet block can represent the interaction between the nodes and edges more efficiently, resulting in a significant performance improvement in rollout prediction settings. However, we still observe mesh explosions after a few frames, as shown in Figure~\ref{fig:big_vegas_visual} and in the supplementary video. 

\vspace{-0.3cm}
\paragraph{MeshGraphNets~\cite{pfaff2020learning}} In concurrent work to us, MeshGraphNets~\cite{pfaff2020learning} were presented for physically-based simulation on a mesh, with an architecture similar to GNS~\cite{sanchez2020learning}. The Lagrangian cloth system presented in their paper is the most closely related approach to our work. Therefore, we followed the input formulation of their example, except that we used the reference mesh to represent the undeformed mesh space as the edge feature. In our implementation, we keep the originally proposed encoders $\epsilon ^M $ and $\epsilon^V$ that embed the edge and node features, but exclude the global (world) feature encoder $\epsilon^W,$ because it is not applicable to our problem setting. Similarly, we kept the MLPs $f^M$ and $f^V,$ but removed the $f^W$  inside the graph block. We used 15 graph blocks in the model, as suggested by their paper. The network has 10 times more parameters than ours; 2,333,187 parameters compared to our 237,571 parameters. Training lasted for 11 days, whereas our network was trained in less than a day.

We report how MeshGraphNets perform on our test character motion sequences in Table~\ref{tab:accuracy_result}. The overall average rollout RMSE of MeshGraphNets is worse than GNS~\cite{sanchez2020learning}. Nevertheless, we note that out of 15 motions, this approach achieved 5 stable rollout predictions without explosions, while GNS~\cite{sanchez2020learning} failed on all of them. Our method outperforms each of the compared methods with respect to the investigated metrics. 

\section{Conclusion}\label{sec:conclusion}

We proposed a \textit{Deep Emulator} for enhancing  skinning-based animations of 3D characters with vivid secondary motion. Our method is inspired by the underlying physical simulation. Specifically, we train a neural network that operates on a local patch of a volumetric simulation mesh of the character, and predicts the updated vertex positions from the current acceleration, velocity, and positions. Being a local method, our network generalizes across 3D character meshes of arbitrary topology.

\begin{wrapfigure}{R}{0.3\linewidth}
\centering
\includegraphics[width=1.0\linewidth]{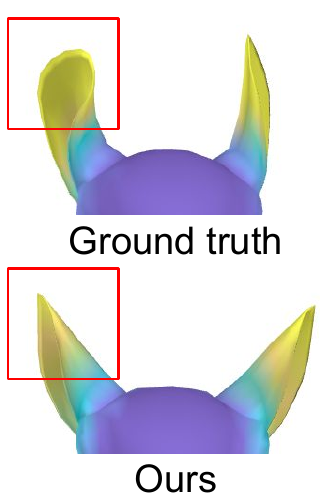}
\end{wrapfigure}
While our method demonstrates plausible secondary dynamics for various 3D characters under complex motions, there are still certain limitations we would like to address in future work. Specifically, we demonstrated that our network trained on a dataset of a volumetric mesh of a sphere can generalize to 3D characters with varying topologies. However, if the local geometric detail of a character is significantly different to those seen during training, e.g., the ears of the mousey character containing many local neighborhood not present in the sphere training data, the quality of our output decreases. One potential avenue for addressing this is to add additional primitive types to training, beyond tetrahedralized spheres. A thorough study on the type of training primitives and motion sequences required to cover the underlying problem domain is an interesting future direction.

\paragraph*{Acknowledgements:}
This research was sponsored in part by
NSF (IIS-1911224), USC Annenberg Fellowship
to Mianlun Zheng, Bosch Research and Adobe Research.

{\small
\bibliographystyle{ieee_fullname}
\bibliography{main}
}

\clearpage
\newpage

\renewcommand{\thepage}{A.\arabic{page}} 
\renewcommand{\thesection}{A.\arabic{section}}  
\renewcommand{\thetable}{A.\arabic{table}}  
\renewcommand{\thefigure}{A.\arabic{figure}}
\setcounter{figure}{0}
\setcounter{table}{0}
\setcounter{section}{0}
\setcounter{page}{1}

\section*{Appendix:}
We sincerely request readers to refer to the link below for more visualization results: \url{https://zhengmianlun.github.io/publications/deepEmulator.html}.
\section{Dataset Information}

In this paper, we trained our network on a sphere dataset but tested it on five character meshes from the Adobe's Mixamo dataset~\cite{mixamo}. Table~\ref{tab:mesh_info} provides detailed information about the five character meshes, including the vertex number and the edge length on the original surface mesh as well as the corresponding uniform volumetric mesh. 



\begin{figure*}[!htbp]
    \centering
    \includegraphics[width=1.0\linewidth]{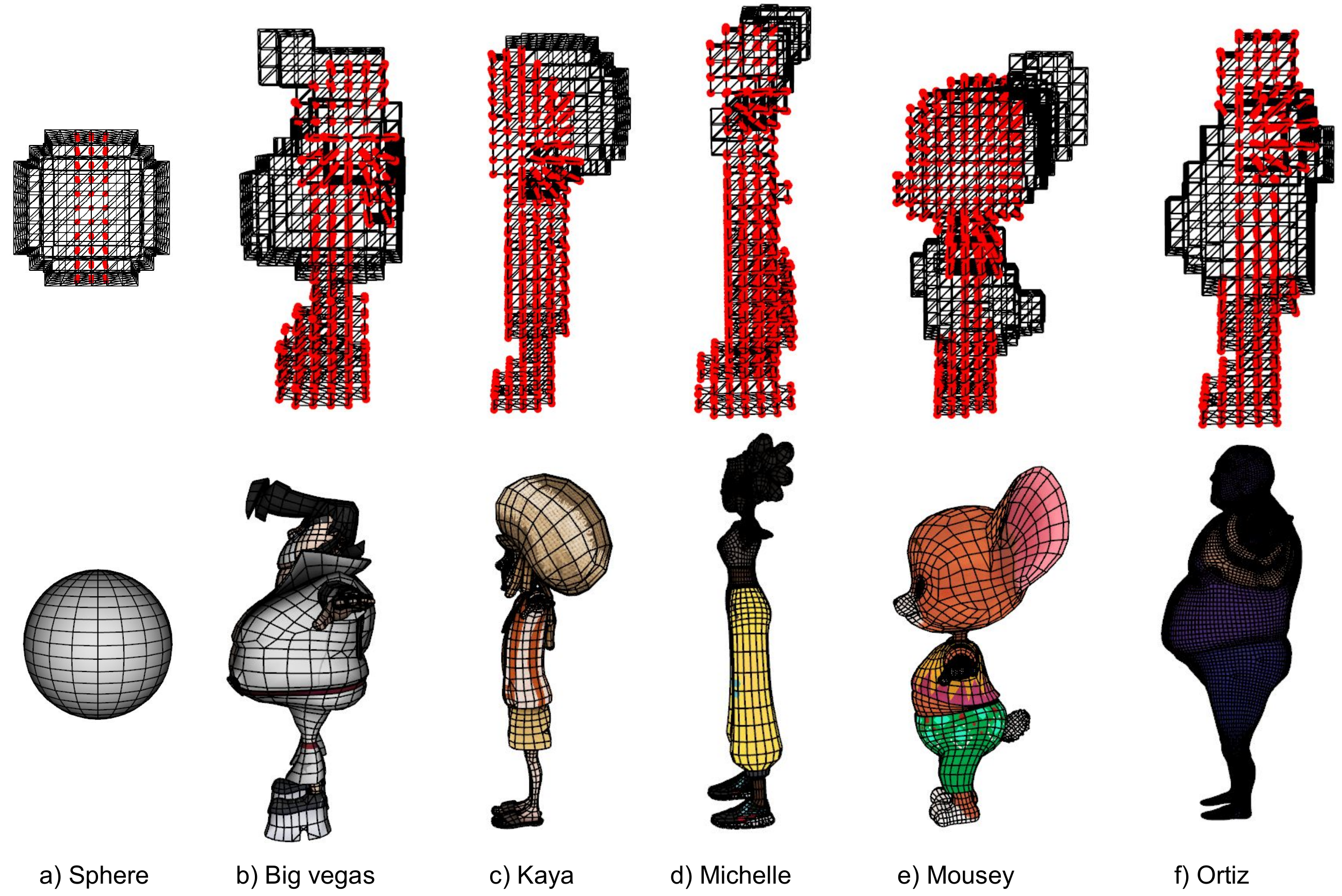}
    \caption{The constraints (red vertices) set on the volumetric mesh surrounding the surface mesh. }
    \label{fig:mesh_info}
\end{figure*}

In Figure~\ref{fig:mesh_info}, we show how we set constraints for each of the meshes, from a side view. The red vertices are constrained to move based on the skinned animation and drive the free vertices to deform with secondary motion.

\section{Full Quantitative and Qualitative Results}

\begin{figure*}[!htbp]
    \centering
    \includegraphics[width=1.0\linewidth]{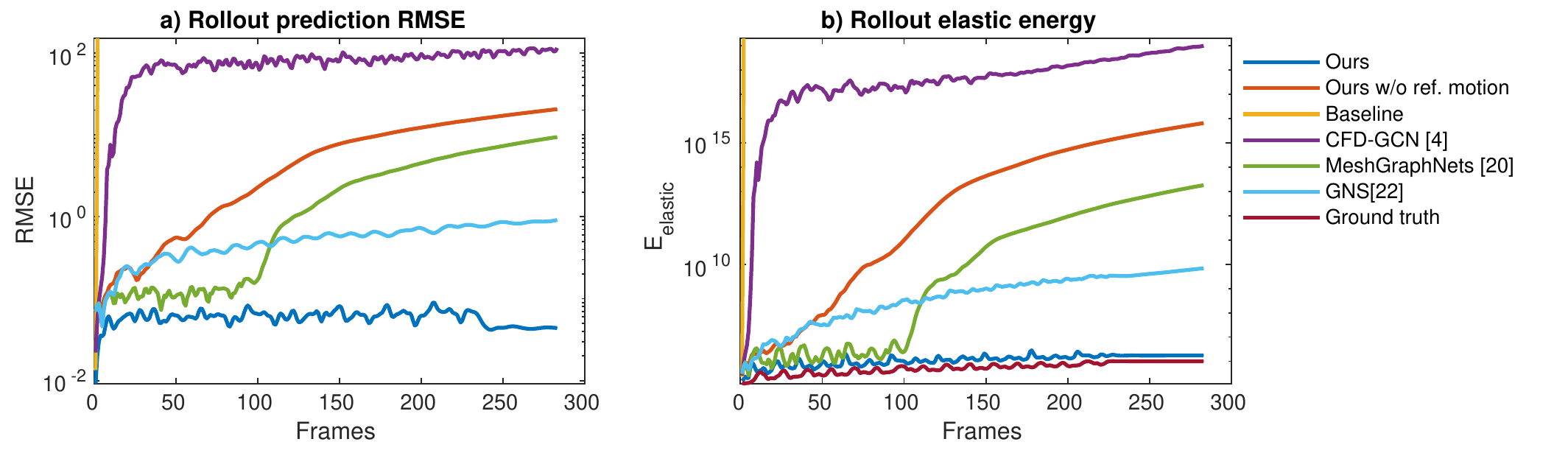}
    \caption{Plot of the quantitative results: Big vegas, 283-frame hip hop dancing 1.}
    \label{fig:big_vegas_1}
\end{figure*}

\begin{figure*}[!htbp]
    \centering
    \includegraphics[width=1.0\linewidth]{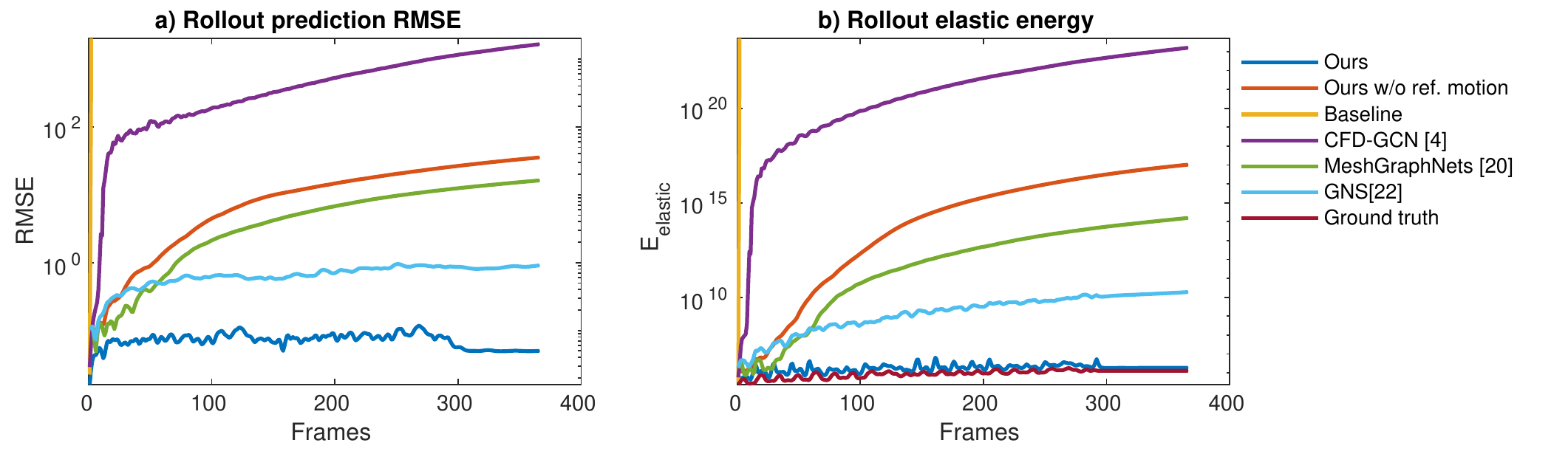}
    \caption{Plot of the quantitative results: Big vegas, 366-frame hip hop dancing 2.}
    \label{fig:big_vegas_2}
\end{figure*}

\begin{figure*}[!htbp]
    \centering
    \includegraphics[width=1.0\linewidth]{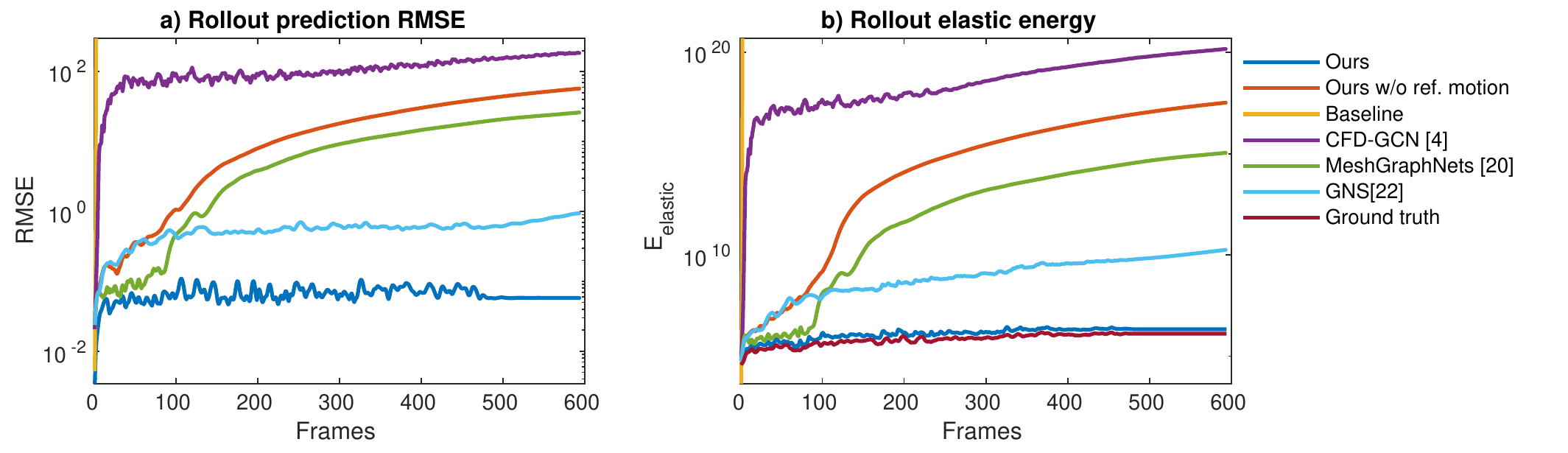}
    \caption{Plot of the quantitative results: Big vegas, 594-frame samba dancing 1.}
    \label{fig:big_vegas_3}
\end{figure*}

\begin{figure*}[!htbp]
    \centering
    \includegraphics[width=1.0\linewidth]{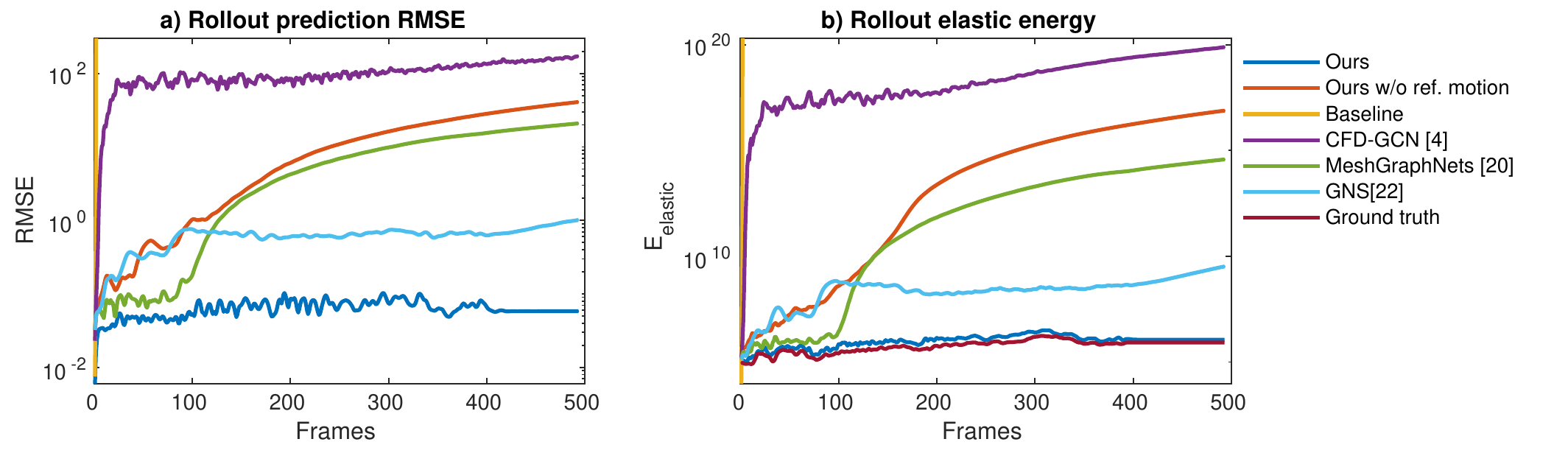}
    \caption{Plot of the quantitative results: Big vegas, 493-frame samba dancing 2.}
    \label{fig:big_vegas_4}
\end{figure*}

\begin{figure*}[!htbp]
    \centering
    \includegraphics[width=1.0\linewidth]{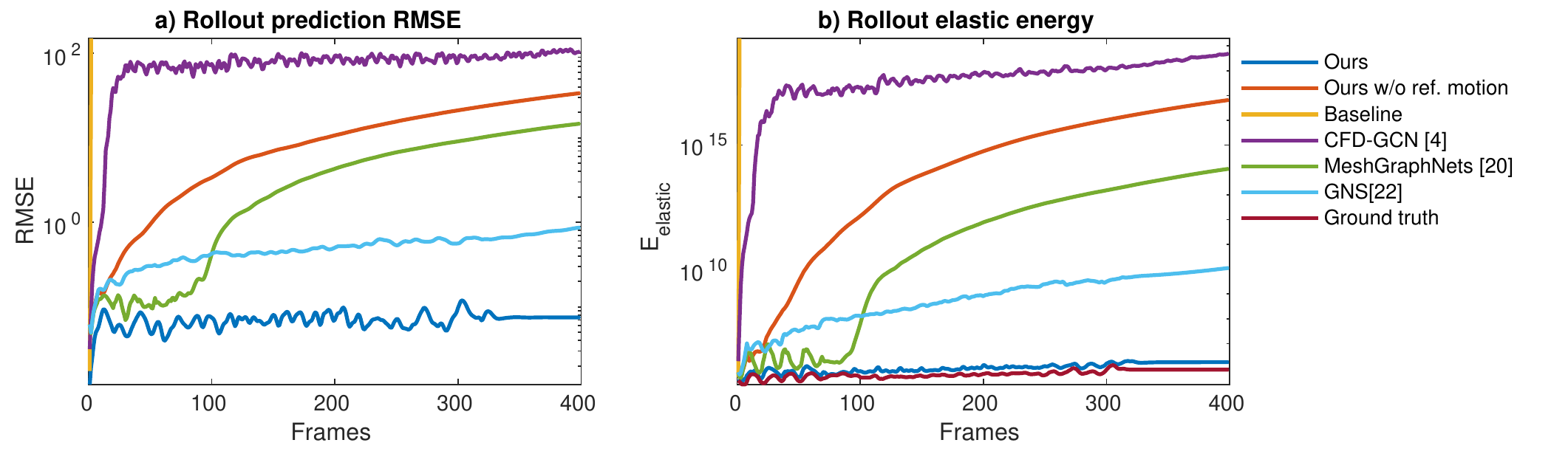}
    \caption{Plot of the quantitative results: Big vegas, 399-frame samba dancing 3.}
    \label{fig:big_vegas_5}
\end{figure*}

\begin{figure*}[!htbp]
    \centering
    \includegraphics[width=1.0\linewidth]{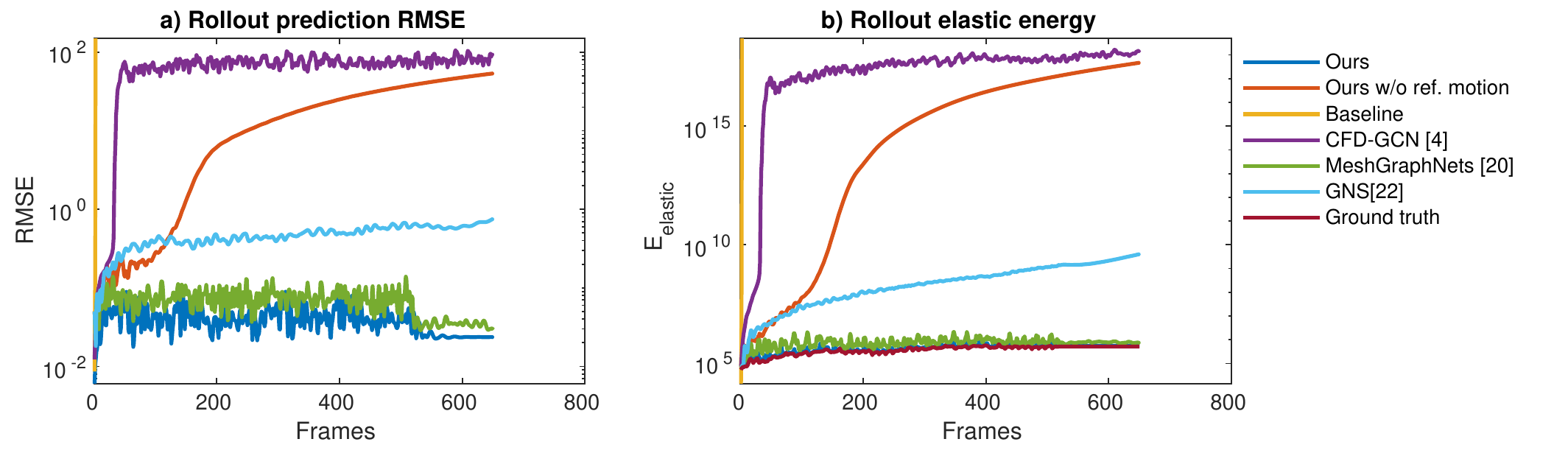}
    \caption{Plot of the quantitative results: Kaya, 650-frame dancing running man.}
    \label{fig:kaya_1}
\end{figure*}

\begin{figure*}[!htbp]
    \centering
    \includegraphics[width=1.0\linewidth]{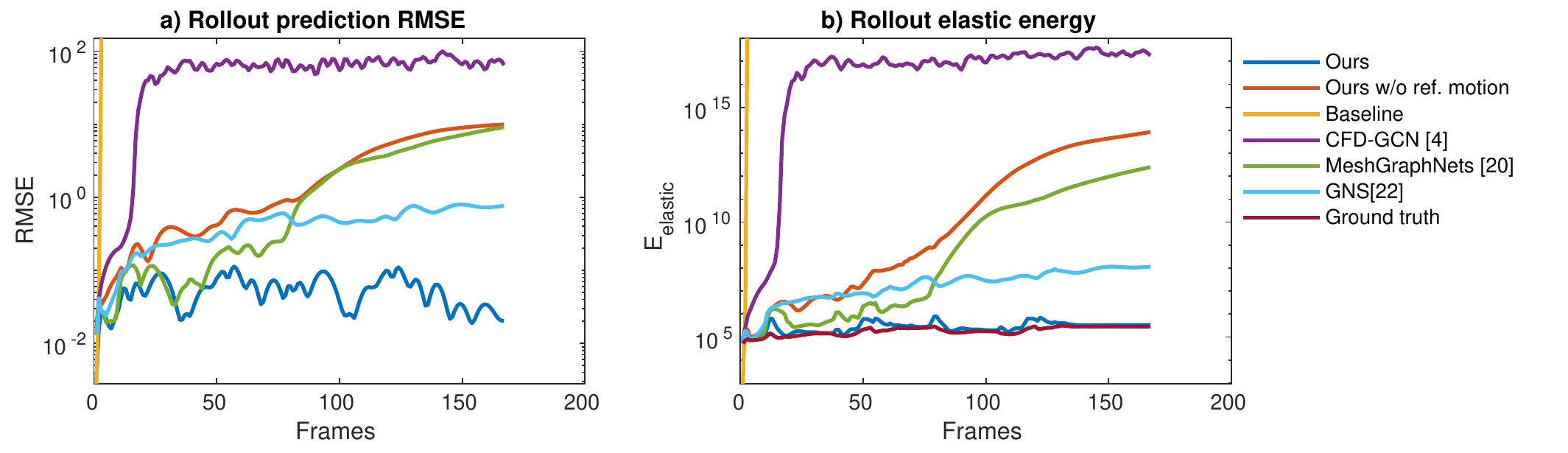}
    \caption{Plot of the quantitative results: Kaya, 167-frame zombie scream.}
    \label{fig:kaya_2}
\end{figure*}

\begin{figure*}[!htbp]
    \centering
    \includegraphics[width=1.0\linewidth]{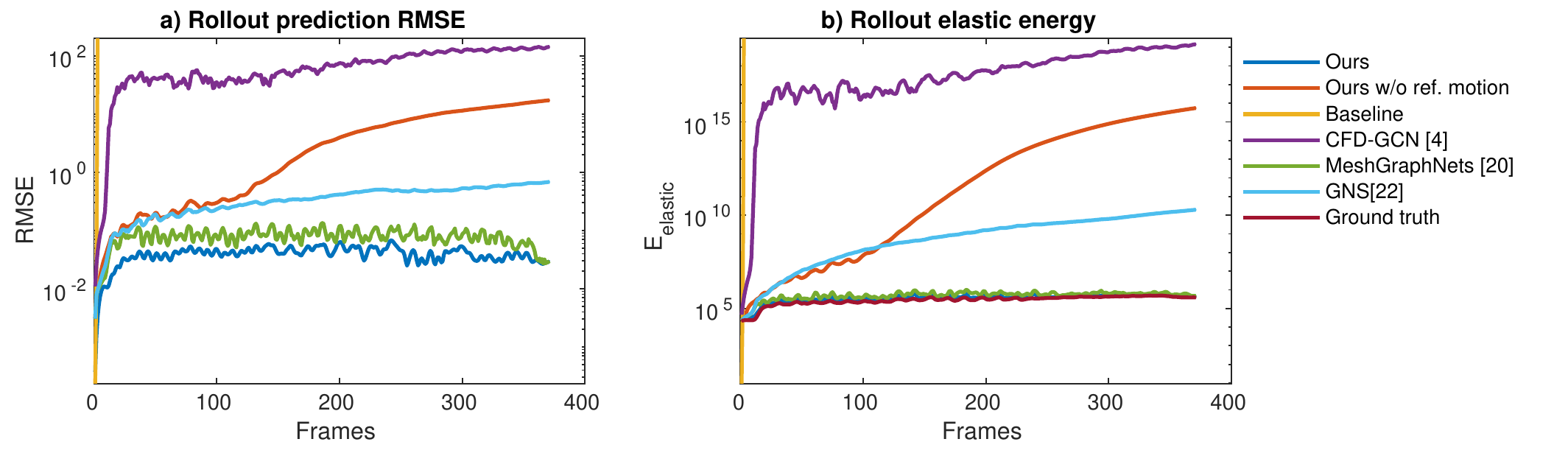}
    \caption{Plot of the quantitative results: Michelle, 371-frame gangnam style.}
    \label{fig:michelle_1}
\end{figure*}

\begin{figure*}[!htbp]
    \centering
    \includegraphics[width=1.0\linewidth]{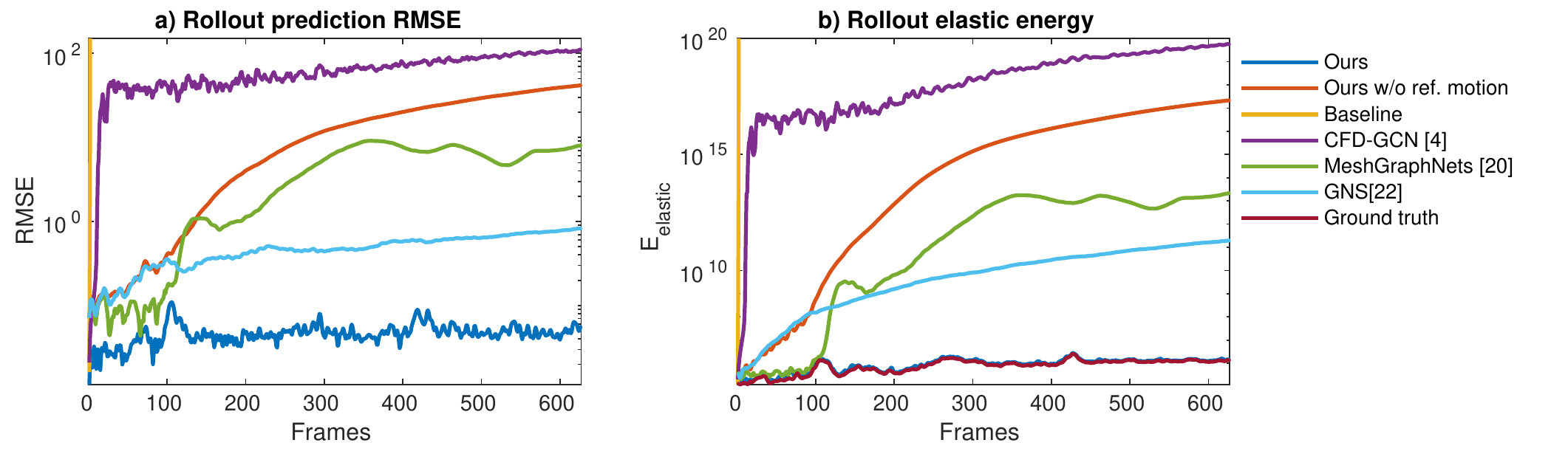}
    \caption{Plot of the quantitative results: Michelle, 627-frame swing dancing 1.}
    \label{fig:michelle_2}
\end{figure*}

\begin{figure*}[!htbp]
    \centering
    \includegraphics[width=1.0\linewidth]{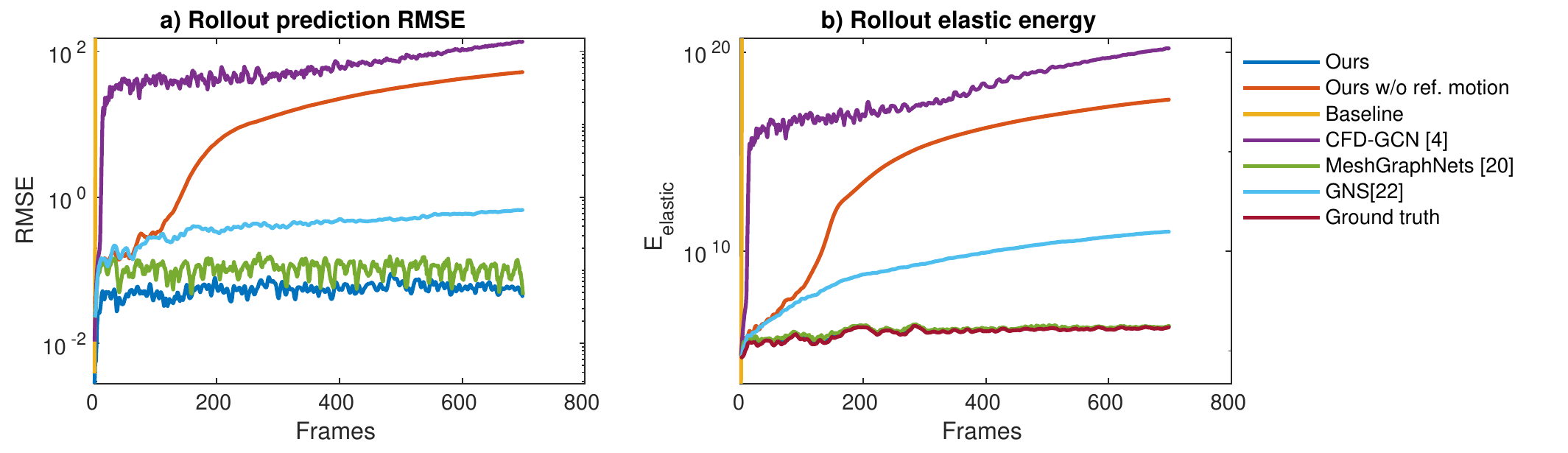}
    \caption{Plot of the quantitative results: Michelle, 699-frame swing dancing 2.}
    \label{fig:michelle_3}
\end{figure*}

\begin{figure*}[!htbp]
    \centering
    \includegraphics[width=1.0\linewidth]{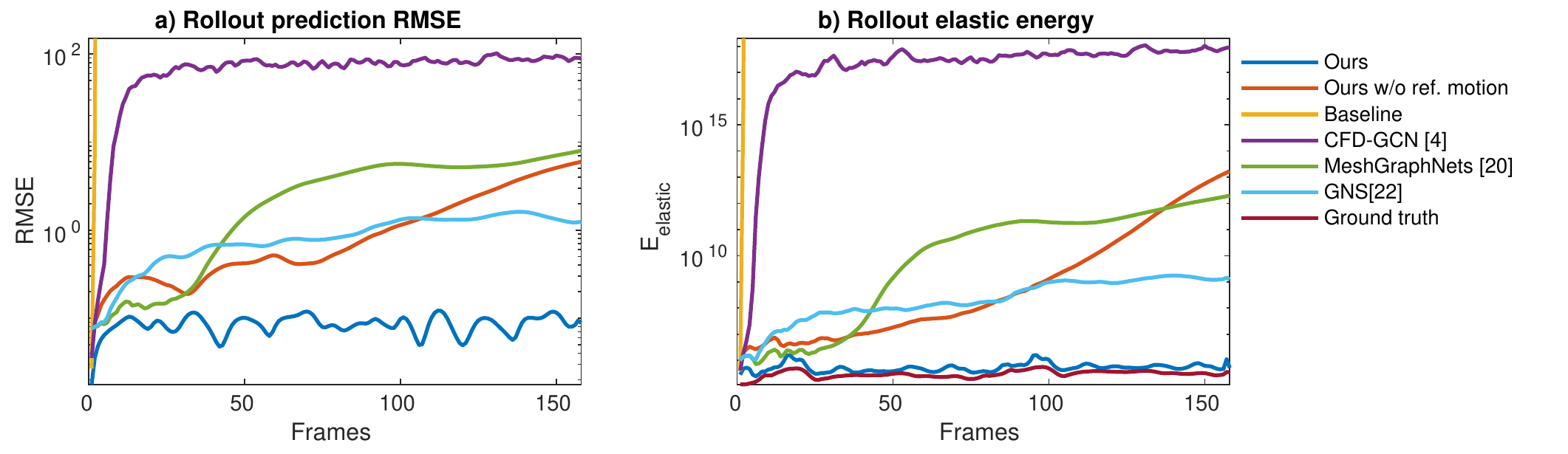}
    \caption{Plot of the quantitative results: Mousey, 158-frame dancing.}
    \label{fig:mousey_1}
\end{figure*}

\begin{figure*}[!htbp]
    \centering
    \includegraphics[width=1.0\linewidth]{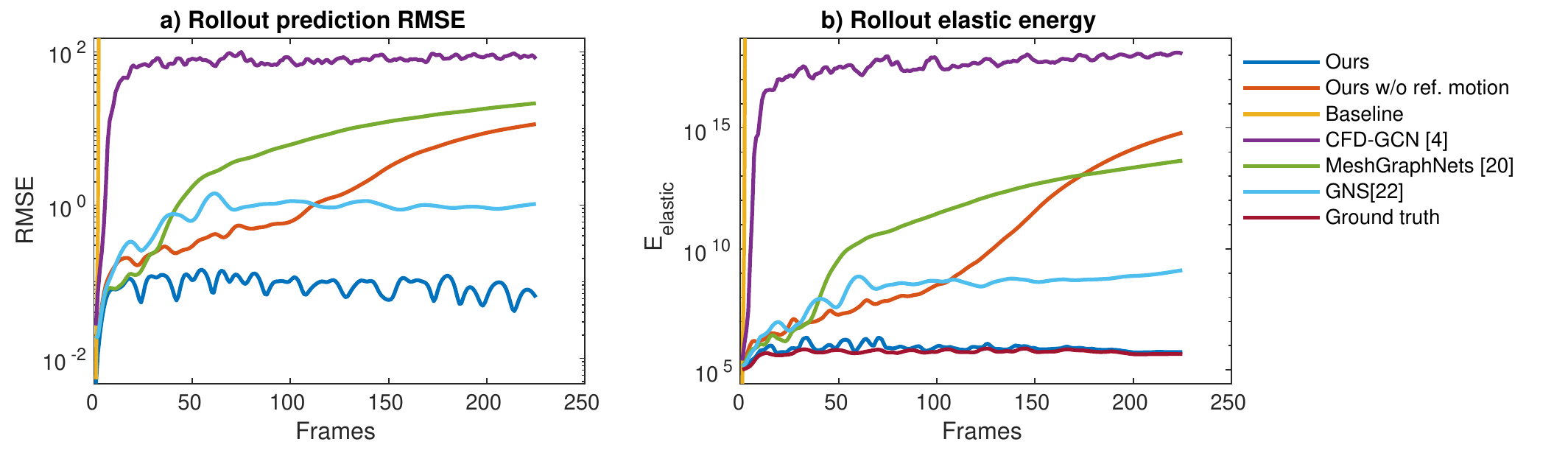}
    \caption{Plot of the quantitative results: Mousey, 255-frame shuffling.}
    \label{fig:mousey_2}
\end{figure*}

\begin{figure*}[!htbp]
    \centering
    \includegraphics[width=1.0\linewidth]{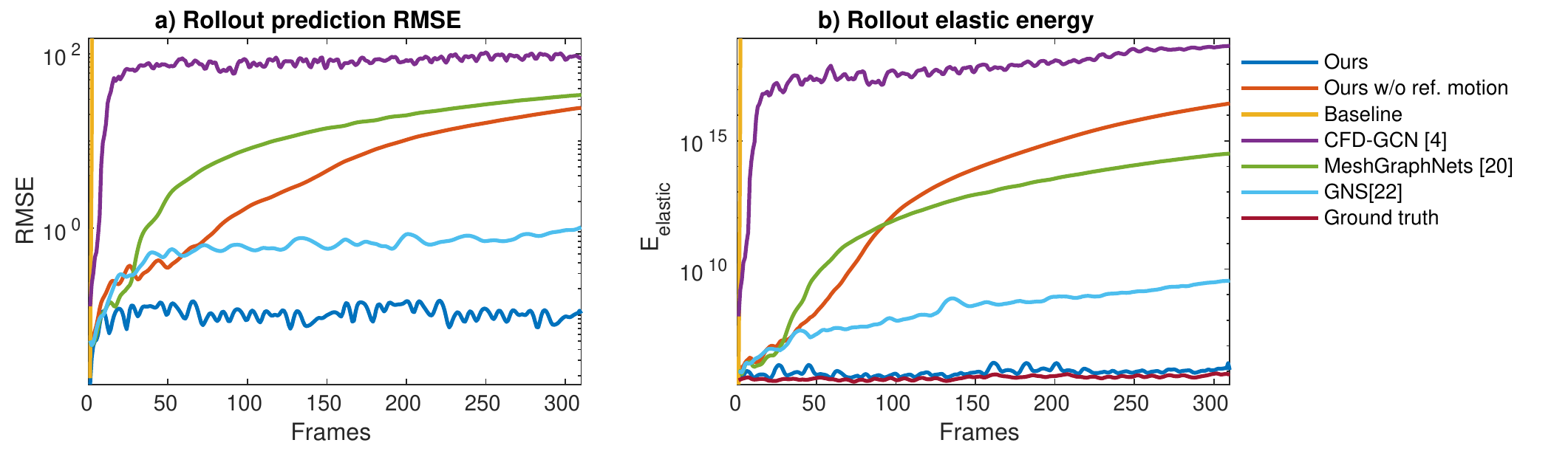}
    \caption{Plot of the quantitative results: Mousey, 627-frame swing dancing.}
    \label{fig:mousey_3}
\end{figure*}

\begin{figure*}[!htbp]
    \centering
    \includegraphics[width=1.0\linewidth]{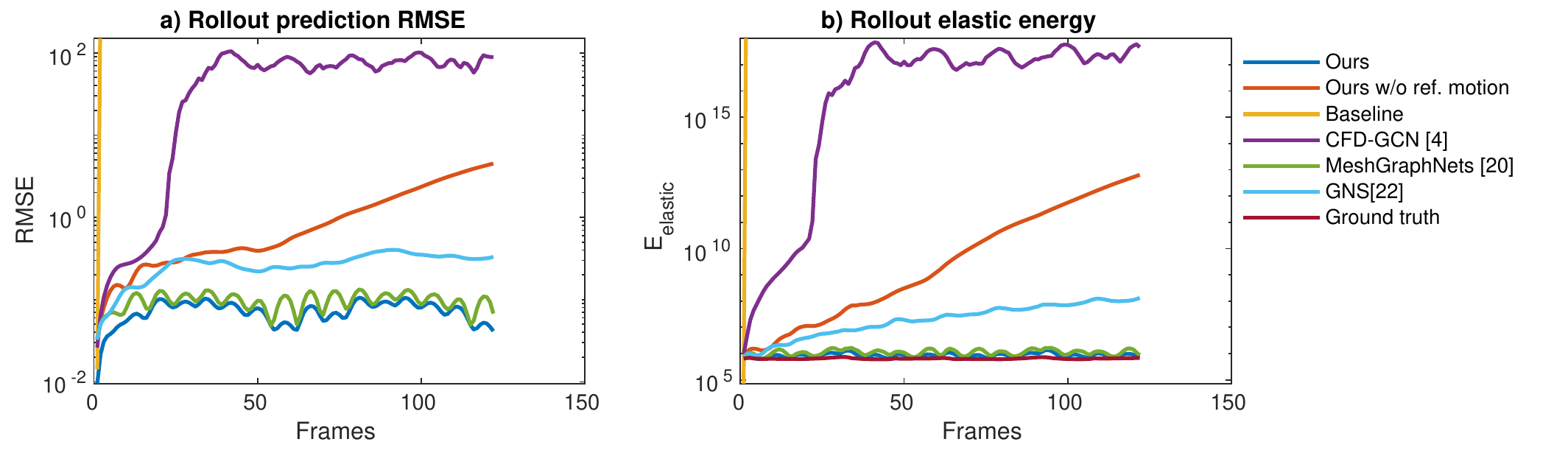}
    \caption{Plot of the quantitative results: Ortiz, 122-frame cross jumps rotation.}
    \label{fig:ortiz_1}
\end{figure*}

\begin{figure*}[!htbp]
    \centering
    \includegraphics[width=1.0\linewidth]{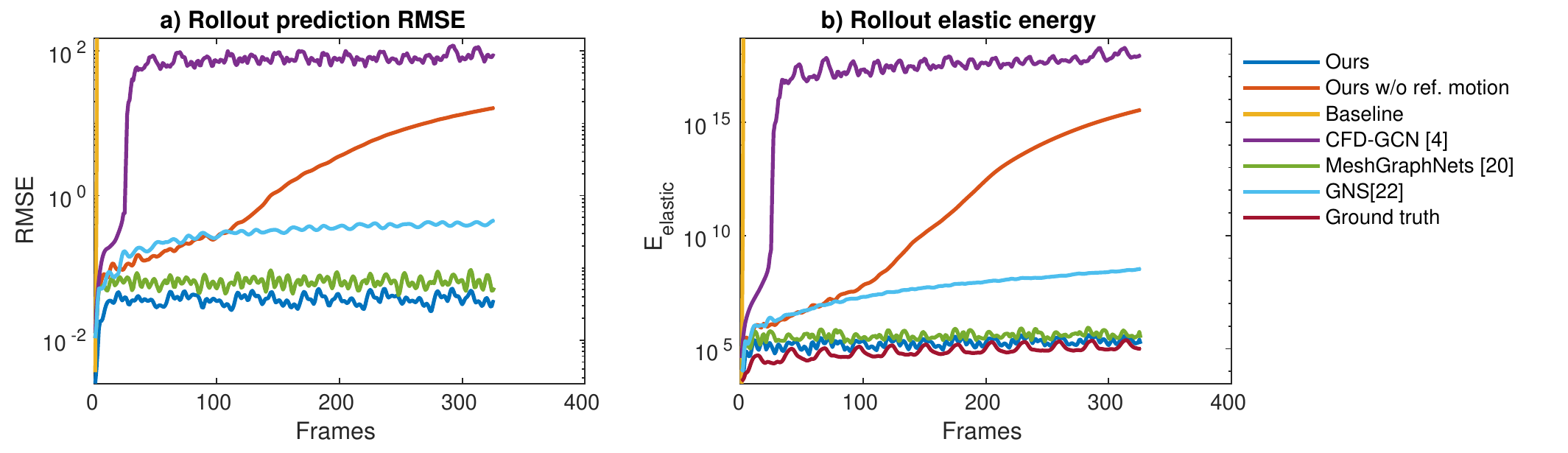}
    \caption{Plot of the quantitative results: Ortiz, 326-frame jazz dancing.}
    \label{fig:ortiz_2}
\end{figure*}

In Tables~\ref{tab:big_vegas_1}-~\ref{tab:ortiz_2}, we provide the quantitative results of our network tested on the five character meshes and 15 motions. The corresponding error plots are given in Figures~\ref{fig:big_vegas_1}-\ref{fig:ortiz_2}. We also provide the error plots for the compared methods. Across all the test cases, our method achieves the most stable rollout prediction with the lowest error. 

In \textbf{DeepEmulator.html}, we provide animation sequences of our results as well as other comparison methods. 

\section{Further Analysis of Baseline Performance}

As introduced in Section~\ref{method_analysis}, we adopted the implicit backward Euler approach (Equation~\ref{eqn:implicit_integration}) as ground truth and the faster explicit central differences integration (Equation~\ref{eqn:explicit_integration}) as the baseline. Although the baseline method is 10 times faster than the implicit integrator with the same time step (1/24 second), it explodes after a few frames. In order to achieve stable simulation results, we found that it requires at least 100 sub-steps ($\Delta t\leq0.0004$). In Table~\ref{tab:time_result_detail}, we provide the per-frame running time of the explicit integration with 50 and 100 steps.

\section{Choice of the Training Dataset}

In Section~\ref{sec:conclusion}, we mentioned a future direction of expanding the training dataset beyond primitive-based datasets such as spheres. Here, we analyze an alternative training dataset, namely the ``Ortiz Dataset'', created by running our physically-based simulator on the volumetric mesh surrounding the Ortiz character (same mesh as in Table~\ref{tab:mesh_info}), with motions acquired from Adobe's Mixamo.  In both datasets, we use the same number of frames. We report our results in Table~\ref{tab:dataset_big_vegas} to~\ref{tab:dataset_ortiz}.

Our experiments show that the network trained on the Sphere Dataset in most cases (75\%) outperforms the Ortiz Dataset. We think there are two reasons for this. First, the local patches in the sphere are general and not specific to any geometry, making the learned neural network more general and therefore more suitable for characters other than Ortiz. Second, the motions in the Ortiz Dataset were created by human artists, and as such these motions follow certain human-selected artistic patterns. The motions in the Sphere Dataset, however, consist of random translations and rotations, which provides a denser sampling of motions in the possible motion space, and therefore improves the robustness of the network.  

\section{Analysis of the Local Patch Size}

In the main paper, we show our network architecture for 1-ring local patches (Figure~\ref{network}). Namely, in the main paper the MLP $f_\beta^\textrm{internal\_force}$ learns to predict the internal forces from the \textbf{1}-ring neighbors around the center vertex. 
Here, we present an ablation study whereby the network learns based on 2-ring local patches, and 3-ring local patches, respectively. For 2-ring local patches, we add an additional MLP $f_{\beta2}^\textrm{internal\_force}$ that receives the inputs from the 2-ring neighbors of the center vertex. The output latent vector is concatenated to the input of the $g_\gamma$ MLP. Similar operation is adopted for the 3-ring local patch network by adding another MLP for the 3-ring internal forces. 

For the training loss, the network achieves the RMSE of 0.00257, 0.00159 and 0.00146 for 1-ring, 2-ring and 3-ring local patches, respectively. In Table~\ref{tab:ring_size_result}, we provide the corresponding test results on the five characters. Overall, we didn't see obvious improvements by increasing the local patch size. 
This could be because 2-ring and 3-ring local patches exhibits larger variability of structure, different to the sphere mesh, particularly for a center vertex close to the boundary.
Therefore, we adopt 1-ring local patches in our paper.

\begin{table*}[!htbp]
    \centering
    \begin{tabular}{l|c|c|c|c|c}
    \hline
     Character & \makecell{Vertex Number \\ (surface mesh)} & \makecell{Edge Length \\ (surface mesh)} & \makecell{Disconnected \\ Components} & \makecell{Vertex Number \\ (tet mesh)} & \makecell{Edge Length \\ (tet mesh)} \\
     \hline
     Big vegas & 3711 & $[0.0024, 0.46]$ & 8 & 1468 & $[0.20, 0.35]$\\
     Kaya & 4260 & $[0.0049, 0.42]$ & 4 & 1417 & $[0.20, 0.35]$\\
     Michelle & 14267 & $[0.00047, 0.27]$ & 1 &  1105 & $[0.20, 0.35]$\\
     Mousey & 6109 & $[0.0023, 0.37]$ & 1 & 2303 & $[0.20, 0.35]$\\
     Ortiz & 24798 & $[0.00087, 0.085]$ & 1 & 1258 & $[0.20, 0.35]$\\
     \hline
    \end{tabular}
    \caption{Detailed information on the five test characters. Each character's surface mesh was re-scaled uniformly to lie exactly within a bounding box of dimensions 5$\times$5$\times$5.}
    \label{tab:mesh_info}
\end{table*}

\begin{table*}[!htbp]
    \centering
    \begin{tabular}{l|c|c|c|c|c}
    \hline
    Methods & Single Frame & Rollout-24 & Rollout-48 & Rollout-All  & \makecell{$E_{elastic}$\\$[min, stdev, max]$} \\
    \hline
    Ground Truth & $\setminus $ & $\setminus $ & $\setminus $ & $\setminus$ & $	[1.21E5, 2.76E5, 1.02E6]	$\\ 	Our Method &	\textbf{0.0098}	&	\textbf{0.053}	&	\textbf{0.063}	&	\textbf{0.059}	& $\mathbf{[1.56E5, 4.92E5, 2.83E6]}$ \\	Ours w/o ref. motion & 	0.058	&	0.19	&	0.55	&	7.70	& $	[2.76E5, 1.5E15, 6.5E15]	$ \\	Baseline & 	$\setminus$	&	8.23E120	&	1.07E121	&	1.57E121	& $	[2.28E5,Nan,4.71E165]	$ \\	CFD-GCN~\cite{de2020combining} & 	0.031	&	50.33	&	72.06	&	79.16	& $	[2.60E5, 2.4E18, 9.9E18]	$ \\	GNS~\cite{sanchez2020learning} & 	0.057	&	0.20	&	0.31	&	0.55	& $	[3.24E5, 1.79E9, 6.83E9]	$ \\	 MeshGraphNets~\cite{pfaff2020learning} & 	0.058	&	0.14	&	0.10	&	2.85	& $	[2.62E5, 3.9E12, 1.8E13]	$ \\			\hline
    \end{tabular}
    \caption{Quantitative results: Big vegas, 283-frame hip hop dancing 1.}
    \label{tab:big_vegas_1}
\end{table*}

\begin{table*}[!htbp]
    \centering
    \begin{tabular}{l|c|c|c|c|c}
    \hline
    Methods & Single Frame & Rollout-24 & Rollout-48 & Rollout-All  & \makecell{$E_{elastic}$\\$[min, stdev, max]$} \\
    \hline
	Ground Truth & $\setminus $ & $\setminus $ & $\setminus $ & $\setminus$ & $	[2.42E5, 3.53E5, 1.77E5]	$\\	Our Method &	\textbf{0.0093}	&	\textbf{0.066}	&	\textbf{0.074}	&	\textbf{0.073}	& $	\mathbf{[3.20E5 ,8.18E5 ,6.32E6]}$ \\	Ours w/o ref. motion & 	0.061	&	0.31	&	0.87	&	14.19	& $	[1.10E6, 2.6E16 ,1.0E17]	$ \\	Baseline & 	$\setminus$	&	7.83E120	&	1.06E121	&	1.79E121	& $	[3.17E5,Nan,4.69E165]	$ \\	CFD-GCN~\cite{de2020combining} & 	0.031	&	67.91	&	110.82	&	591.46	& $	[5.79E5 ,3.8E22 ,1.6E23]	$ \\	GNS~\cite{sanchez2020learning} & 	0.060	&	0.32	&	0.50	&	0.68	& $	[1.90E6 ,5.64E9 ,1.9E10]	$ \\	 MeshGraphNets~\cite{pfaff2020learning} & 	0.062	&	0.16	&	0.38	&	6.58	& $	[7.58E5, 3.9E13 ,1.6E14]	$ \\			\hline
    \end{tabular}
    \label{tab:big_vegas_2}
    \caption{Quantitative results: Big vegas, 366-frame hip hop dancing 2.}
\end{table*}

\begin{table*}[!htbp]    
    \centering
    \begin{tabular}{l|c|c|c|c|c}
    \hline
    Methods & Single Frame & Rollout-24 & Rollout-48 & Rollout-All  & \makecell{$E_{elastic}$\\$[min, stdev, max]$} \\
    \hline
	Ground Truth & $\setminus $ & $\setminus $ & $\setminus $ & $\setminus$ & $	[3.82E4 ,3.99E5,1.50E6]	$\\	Our Method &	\textbf{0.0062}	&	\textbf{0.050}	&	\textbf{0.057}	&	\textbf{0.065}	& $\mathbf{[4.12E4, 6.68E5, 2.76E6]}	$ \\	Ours w/o ref. motion & 	0.047	&	0.15	&	0.36	&	21.37	& $	[6.19E4, 8.1E16 ,3.4E17]	$ \\	Baseline & 	$\setminus$	&	7.67E120	&	1.07E121	&	2.51E121	& $	[4.40E3,Nan,4.68E165]	$ \\	CFD-GCN~\cite{de2020combining} & 	0.027	&	47.90	&	76.76	&	110.26	& $	[8.89E4, 3.8E17 ,1.5E19] $ \\	GNS~\cite{sanchez2020learning} & 	0.046	&	0.17	&	0.32	&	0.56	& $	[5.77E4, 4.20E9, 1.8E10]	$ \\	 MeshGraphNets~\cite{pfaff2020learning} & 	0.048	&	0.084	&	0.081	&	10.12	& $	[6.23E4, 2.8E14, 1.1E15]	$ \\	\hline
    \end{tabular}
    \label{tab:big_vegas_3}
    \caption{Quantitative results: Big vegas, 594-frame samba dancing 1.}
\end{table*}

\begin{table*}[!htbp]
    \centering
    \begin{tabular}{l|c|c|c|c|c}
    \hline
    Methods & Single Frame & Rollout-24 & Rollout-48 & Rollout-All  & \makecell{$E_{elastic}$\\$[min, stdev, max]$} \\
    \hline
	Ground Truth & $\setminus $ & $\setminus $ & $\setminus $ & $\setminus$ & $	[8.02E4 ,3.72E5,1.74E6]	$\\	Our Method &	\textbf{0.0058}	&	\textbf{0.052}	&	\textbf{0.050}	&	\textbf{0.064}	& $\mathbf{[9.76E4 ,6.35E5 ,3.18E6]}$ \\	Ours w/o ref. motion & 	0.038	&	0.13	&	0.40	&	13.88	& $	[1.45E5, 1.8E16 ,7.7E16]	$ \\	Baseline & 	$\setminus$	&	7.67E120	&	9.97E120	&	2.15E121	& $	[9.58E3,Nan,4.70E165]	$ \\	CFD-GCN~\cite{de2020combining} & 	0.027	&	75.12	&	86.49	&	101.63	& $	[1.59E5, 1.9E19 ,7.7E19] $ \\	GNS~\cite{sanchez2020learning} & 	0.037	&	0.17	&	0.30	&	0.62	& $	[1.29E5, 5.85E8, 3.32E9]	$ \\	 MeshGraphNets~\cite{pfaff2020learning} & 	0.040	&	0.090	&	0.091	&	7.89	& $	[1.36E5, 9.3E13, 3.8E14]	$ \\		\hline
    \end{tabular}
    \label{tab:big_vegas_4}
    \caption{Quantitative results: Big vegas, 493-frame samba dancing 2.}
\end{table*}

\begin{table*}[!htbp]
    \centering
    \begin{tabular}{l|c|c|c|c|c}
    \hline
    Methods & Single Frame & Rollout-24 & Rollout-48 & Rollout-All  & \makecell{$E_{elastic}$\\$[min, stdev, max]$} \\
    \hline
	Ground Truth & $\setminus $ & $\setminus $ & $\setminus $ & $\setminus$ & $	[2.24E5, 2.06E5, 1.34E6]	$\\	Our Method &	\textbf{0.0065} &	\textbf{0.062}	&	\textbf{0.054}	&	\textbf{0.070}	& $\mathbf{[2.92E5, 4.31E5, 1.98E6]}$ \\	Ours w/o ref. motion & 	0.040	&	0.35	&	0.90	&	12.82	& $	[5.12E5, 1.6E16 ,6.6E16]	$ \\	Baseline & 	$\setminus$	&	7.76E120	&	1.02E121	&	1.96E121	& $	[4.48E5,Nan,4.70E165]	$ \\	CFD-GCN~\cite{de2020combining} & 	0.083	&	50.17	&	71.88	&	79.87	& $	[1.95E6, 1.1E17 ,4.9E17]	$ \\	GNS~\cite{sanchez2020learning} & 	0.040	&	0.18	&	0.31	&	0.50	& $	[5.61E5, 2.85E9, 1.1E10]	$ \\	 MeshGraphNets~\cite{pfaff2020learning} & 	0.043	&	0.13	&	0.11	&	5.24	& $	[4.09E5, 2.6E13, 1.1E15]	$ \\
    \hline
    \end{tabular}
    \caption{Quantitative results: Big vegas, 399-frame samba dancing 3.}
    \label{tab:big_vegas_5}
\end{table*}
		
\begin{table*}[!htbp]
    \centering
    \begin{tabular}{l|c|c|c|c|c}
    \hline
    Methods & Single Frame & Rollout-24 & Rollout-48 & Rollout-All  & \makecell{$E_{elastic}$\\$[min, stdev, max]$} \\
    \hline
    Ground Truth & $\setminus $ & $\setminus $ & $\setminus $ & $\setminus$ & $	[5.47E4, 1.31E5, 6.40E51]	$\\	Our Method & \textbf{0.0067} & \textbf{0.054} &	\textbf{0.058} & \textbf{0.041} & $\mathbf{[5.68E4, 1.50E5, 1.03E6]}$ \\	Ours w/o ref. motion & 	0.042	&	0.097	&	0.15	&	20.02	& $	[7.70E4, 1.1E16 ,4.7E17]	$ \\	Baseline & 	$\setminus$	&	6.21E120	&	8.00E120	&	2.00E121	& $	[1.41E4,Nan,3.15E165]	$ \\	CFD-GCN~\cite{de2020combining} & 	0.016	&	0.22	&	72.15	&	69.87	& $	[7.33E4, 3.6E17 ,1.7E18]	$ \\	GNS~\cite{sanchez2020learning} & 	0.041	&	0.15	&	0.28	&	0.47	& $	[7.20E4, 8.63E8, 4.02E9]	$ \\	 MeshGraphNets~\cite{pfaff2020learning} & 	0.042	&	0.063	&	0.084	&	0.068	& $	[7.27E4,3.03E5,2.12E6]	$ \\
    \hline
    \end{tabular}
    \label{tab:kaya_1}
    \caption{Quantitative results: Kaya, 650-frame dancing running man.}
\end{table*}

\begin{table*}[!htbp]
    \centering
    \begin{tabular}{l|c|c|c|c|c}
    \hline
    Methods & Single Frame & Rollout-24 & Rollout-48 & Rollout-All  & \makecell{$E_{elastic}$\\$[min, stdev, max]$} \\
    \hline
    Ground truth & $\setminus $ & $\setminus $ & $\setminus $ & $\setminus$ & $	[5.14E4, 7.09E4, 3.15E5]	$\\	Our Method &	\textbf{0.0075}	&	\textbf{0.083}	&	\textbf{0.067}	&	\textbf{0.054}	& $	\mathbf{[5.17E4, 1.49E5, 7.76E5]}	$ \\	Ours w/o ref. motion & 	0.030	&	0.23	&	0.41	&	3.16	& $	[5.87E4, 2.1E13, 8.4E13]	$ \\	Baseline & 	$\setminus$	&	5.66E120	&	7.98E120	&	8.99E120	& $	[9.30E2,Nan,3.15E165]	$ \\	CFD-GCN~\cite{de2020combining} & 	0.023	&	35.34 &	62.31	&	59.14	& $	[7.21E4, 8.9E16 ,4.0E17]	$ \\	GNS~\cite{sanchez2020learning} & 	0.029	&	0.21	&	0.29	&	0.45	& $	[5.83E4, 3.65E7, 1.16E8]	$ \\	 MeshGraphNets~\cite{pfaff2020learning} & 	0.03	&	0.11	&	0.15	&	2.44	& $	[5.88E4, 5.3E10, 2.5E12]	$ \\
    \hline
    \end{tabular}
    \caption{Quantitative results: Kaya, 167-frame zombie scream.}
    \label{tab:kaya_2}
\end{table*}

\begin{table*}[!htbp]
    \centering
    \begin{tabular}{l|c|c|c|c|c}
    \hline
    Methods & Single Frame & Rollout-24 & Rollout-48 & Rollout-All  & \makecell{$E_{elastic}$\\$[min, stdev, max]$} \\
    \hline
    Ground truth & $\setminus $ & $\setminus $ & $\setminus $ & $\setminus$ & $	[2.18E4, 1.12E5, 4.93E5]	$\\	Our Method & \textbf{0.0041} &	\textbf{0.033}	&	\textbf{0.033}	&	\textbf{0.04}	& $	\mathbf{[2.20E4, 1.25E5, 6.45E5]}$ \\	Ours w/o ref. motion & 	0.060	&	0.12	&	0.19	&	5.24	& $	[2.27E4, 1.2E15 ,5.5E15]	$ \\	Baseline & 	$\setminus$	&	5.81E120	&	7.86E120	&	1.52E121	& $	[9.26E0,Nan,1.27E165]	$ \\	CFD-GCN~\cite{de2020combining} & 	0.017	&	27.36	&	42.25	&	73.72	& $	[4.58E4, 3.7E18 ,1.4E19]	$ \\	GNS~\cite{sanchez2020learning} & 	0.060	&	0.10	&	0.19	&	0.37	& $	[2.26E4, 4.79E9, 1.9E10]	$ \\	 MeshGraphNets~\cite{pfaff2020learning} & 	0.06	&	0.082	&	0.11	&	0.079	& $	[2.26E4, 1.76E5, 9.94E5]	$ \\
    \hline
    \end{tabular}
    \label{tab:michelle_1}
    \caption{Quantitative results: Michelle, 371-frame gangnam style.}
\end{table*}

\begin{table*}[!htbp]
    \centering
    \begin{tabular}{l|c|c|c|c|c}
    \hline
    Methods & Single Frame & Rollout-24 & Rollout-48 & Rollout-All  & \makecell{$E_{elastic}$\\$[min, stdev, max]$} \\
    \hline
    Ground truth & $\setminus $ & $\setminus $ & $\setminus $ & $\setminus$ & $	[1.22836E5, 4.67E5, 2.60E6]	$\\	Our Method &	\textbf{0.0056}	&	\textbf{0.025}	&	\textbf{0.024}	&	\textbf{0.047}	& $\mathbf{[1.43E5 ,5.06E5,2.79E6]}$ \\	Ours w/o ref. motion & 	0.082	&	0.13	&	0.15	&	15.20	& $	[2.96E5, 5.3E16 ,2.2E17]	$ \\	Baseline & 	$\setminus$	&	6.25E120	&	7.47E120	&	2.25E121	& $	[1.25E5,Nan,1.27E165]	$ \\	CFD-GCN~\cite{de2020combining} & 	0.019	&	36.06	&	36.80	&	64.16	& $	[2.05E5, 1.5E19 ,5.7E19]	$ \\	GNS~\cite{sanchez2020learning} & 	0.082	&	0.12	&	0.14	&	0.48	& $	[2.35E5, 5.1E10, 2.0E11]	$ \\	 MeshGraphNets~\cite{pfaff2020learning} & 	0.082	&	0.065	&	0.077	&	4.31	& $	[2.06E5, 6.5E12, 2.2E13]	$ \\
    \hline
    \end{tabular}
    \label{tab:michelle_2}
    \caption{Quantitative results: Michelle, 627-frame swing dancing 1.}
\end{table*}

\begin{table*}[!htbp]
    \centering
    \begin{tabular}{l|c|c|c|c|c}
    \hline
    Methods & Single Frame & Rollout-24 & Rollout-48 & Rollout-All  & \makecell{$E_{elastic}$\\$[min, stdev, max]$} \\
    \hline
    Ground truth & $\setminus $ & $\setminus $ & $\setminus $ & $\setminus$ & $	[4.46E4,4.18E5,1.82E6]	$\\	Our Method &	\textbf{0.0056}	&	\textbf{0.049}	&	\textbf{0.037}	&	\textbf{0.055}	& $\mathbf{[4.54E4,4.48E5,2.01E6]	}$ \\	Ours w/o ref. motion & 	0.086	&	0.14	&	0.16	&	20.14	& $	[5.66E4,1.1E17 ,4.2E17]	$ \\	Baseline & 	$\setminus$	&	5.71E120	&	7.69E120	&	2.35E121	& $	[2.28E3,Nan,1.27E165]	$ \\	CFD-GCN~\cite{de2020combining} & 	0.019	&	25.05	&	42.43	&	64.23	& $	[5.58E4 ,3.6E18 ,1.6E19]	$ \\	GNS~\cite{sanchez2020learning} & 	0.085	&	0.13	&	0.19	&	0.43	& $	[5.31E4,2.7E10,1.0E14]	$ \\	 MeshGraphNets~\cite{pfaff2020learning} & 	0.086	&	0.11	&	0.094	&	0.11	& $	[5.59E4,4.61E5,2.20E6]	$ \\			
    \hline
    \end{tabular}
    \caption{Quantitative results: Michelle, 699-frame swing dancing 2.}
    \label{tab:michelle_3}
\end{table*}

\begin{table*}[!htbp]
    \centering
    \begin{tabular}{l|c|c|c|c|c}
    \hline
    Methods & Single Frame & Rollout-24 & Rollout-48 & Rollout-All  & \makecell{$E_{elastic}$\\$[min, stdev, max]$} \\
    \hline
    Ground truth & $\setminus $ & $\setminus $ & $\setminus $ & $\setminus$ & $	[1.20E5,8.97E4,5.66E5]	$\\	Our Method &	\textbf{0.0080}	&	\textbf{0.077}	&	\textbf{0.10}	&	\textbf{0.086}	& $\mathbf{[2.45E5 ,2.49E5,1.60E6]}$ \\	Ours w/o ref. motion & 	0.057	&	0.24	&	0.42	&	1.43	& $	[1.27E6,2.7E11, 1.7E13]	$ \\	Baseline & 	$\setminus$	&	7.78E120	&	1.07E121	&	1.24E121	& $	[3.79E5,Nan,7.19E165]	$ \\	CFD-GCN~\cite{de2020combining} & 	0.041	&	56.37	&	82.57	&	72.71	& $	[4.10E5, 2.5E17 ,1.1E18]	$ \\	GNS~\cite{sanchez2020learning} & 	0.057	&	0.50	&	0.69	&	0.92	& $	[1.02E6,5.74E8 ,1.76E9]	$ \\	 MeshGraphNets~\cite{pfaff2020learning} & 	0.057	&	0.15	&	1.30	&	3.48	& $	[7.24E5,4.3E10,2.0E15]	$ \\
    \hline
    \end{tabular}
    \label{tab:mousey_1}
    \caption{Quantitative results: Mousey, 158-frame dancing.}
\end{table*}

\begin{table*}[!htbp]
    \centering
    \begin{tabular}{l|c|c|c|c|c}
    \hline
    Methods & Single Frame & Rollout-24 & Rollout-48 & Rollout-All  & \makecell{$E_{elastic}$\\$[min, stdev, max]$} \\
    \hline
    Ground truth & $\setminus $ & $\setminus $ & $\setminus $ & $\setminus$ & $	[1.01E5,1.14E5,7.62E5]	$\\	Our Method &	\textbf{0.0066}	&	\textbf{0.067}	&	\textbf{0.11}	&	\textbf{0.09}	& $	\mathbf{[1.04E5,3.37E5, 2.12E6]}	$ \\	Ours w/o ref. motion & 	0.036	&	0.19	&	0.28	&	2.95 & $	[1.43E5,1.2E13 ,6.3E14]	$ \\	Baseline & 	$\setminus$	&	8.29E120	&	1.12E121	&	1.51E121	& $	[2.69E4,Nan,7.23E165]	$ \\	CFD-GCN~\cite{de2020combining} & 	0.043	&	68.29	&	78.87	&	75.08	& $	[2.42E5, 3.4E17 ,1.3E18]	$ \\	GNS~\cite{sanchez2020learning} & 	0.036	&	0.26	&	0.62	&	0.86	& $	[1.37E5, 2.92E8, 1.31E9]	$ \\	 MeshGraphNets~\cite{pfaff2020learning} & 	0.037	&	0.15	&	1.61	&	8.63	& $	[1.39E5,1.1E15 ,4.4E15]	$ \\	
    \hline
    \end{tabular}
    \label{tab:mousey_2}
    \caption{Quantitative results: Mousey, 255-frame shuffling.}
\end{table*}
 
\begin{table*}[!htbp]
    \centering
    \begin{tabular}{l|c|c|c|c|c}
    \hline
    Methods & Single Frame & Rollout-24 & Rollout-48 & Rollout-All  & \makecell{$E_{elastic}$\\$[min, stdev, max]$} \\
    \hline
    Ground truth & $\setminus $ & $\setminus $ & $\setminus $ & $\setminus$ & $	[4.00E5,9.80E4,8.39E5]	$\\	Our Method & \textbf{0.0090}	&	\textbf{0.087}	&	\textbf{0.10}	&	\textbf{0.10}	& $\mathbf{[5.27E5,3.23E5, 2.19E6]}$ \\	Ours w/o ref. motion & 	0.039	&	0.36	&	0.35	&	7.86	& $	[1.23E6,6.7E15 ,2.9E16]	$ \\	Baseline & 	$\setminus$	&	8.48E120	&	1.10E121	&	1.81E121	& $	[3.12E5,Nan,7.20E165]	$ \\	CFD-GCN~\cite{de2020combining} & 	0.17	&	66.19	&	73.93	&	78.95	& $	[1.42E8,1.5E17 ,5.2E17]	$ \\	GNS~\cite{sanchez2020learning} & 	0.039	&	0.28	&	0.50	&	0.63	& $	[9.23E5,8.74E8, 3.58E9]	$ \\	 MeshGraphNets~\cite{pfaff2020learning} & 	0.040	&	0.21	&	2.14	&	14.97	& $	[7.86E5,8.4E13 ,3.2E14]	$ \\
    \hline
    \end{tabular}
    \caption{Quantitative results: Mousey, 627-frame swing dancing.}
    \label{tab:mousey_3}
\end{table*}

\begin{table*}[!htbp]
    \centering
    \begin{tabular}{l|c|c|c|c|c}
    \hline
    Methods & Single Frame & Rollout-24 & Rollout-48 & Rollout-All  & \makecell{$E_{elastic}$\\$[min, stdev, max]$} \\
    \hline
	Ground truth & $\setminus $ & $\setminus $ & $\setminus $ & $\setminus$ & $	[6.02E5,3.24E4,7.38E5]	$\\	Our Method &	\textbf{0.0057}	&	\textbf{0.082}	&	\textbf{0.077}	&	\textbf{0.073}	& $\mathbf{[6.16E5,1.55E5,1.35E6]}$ \\	Ours w/o ref. motion & 	0.041	&	0.29	&	0.40	&	1.20	& $	[8.01E5,3.6E12,2.1E13]	$ \\	Baseline & 	$\setminus$	&	8.15E120	&	1.09E121	&	1.09E121	& $	[7.17E4,Nan,4.00E165]	$ \\	CFD-GCN~\cite{de2020combining} & 	0.030	&	10.71	&	65.43	&	58.90	& $	[9.47E5,1.7E18 ,7.0E18]	$ \\	GNS~\cite{sanchez2020learning} & 	0.040	&	0.30	&	0.22	&	0.27	& $	[7.46E5,3.70E7,1.31E8]	$ \\	 MeshGraphNets~\cite{pfaff2020learning} & 	0.042	&	0.090	&	0.088	&	0.096	& $	[6.35E5,2.67E5,1.66E6]	$ \\
    \hline
    \end{tabular}
    \label{tab:ortiz_1}
    \caption{Quantitative results: Ortiz, 122-frame cross jumps rotation.}
\end{table*}

\begin{table*}[!htbp]
    \centering
    \begin{tabular}{l|c|c|c|c|c}
    \hline
    Methods & Single Frame & Rollout-24 & Rollout-48 & Rollout-All  & \makecell{$E_{elastic}$\\$[min, stdev, max]$} \\
    \hline
    Ground truth & $\setminus $ & $\setminus $ & $\setminus $ & $\setminus$ & $	[3.81E3, 4.86E4, 2.40E5]	$\\	Our Method &	\textbf{0.0039}	&	\textbf{0.039}	&	\textbf{0.032}	&	\textbf{0.036}	& $	\mathbf{[4.84E3, 7.56E4, 4.02E5]}$ \\	Ours w/o ref. motion & 	0.034	&	0.10	&	0.17	&	4.18	& $	[1.62E4, 1.7E15 ,7.9E15]	$ \\	Baseline & 	$\setminus$	&	7.40E120	&	9.48E120	&	1.62E121	& $	[2.89E3,Nan,3.99E165]	$ \\	CFD-GCN~\cite{de2020combining} & 	0.017	&	0.57	&	83.52	&	71.93	& $	[3.96E4, 3.4E17, 1.9E18]	$ \\	GNS~\cite{sanchez2020learning} & 	0.034	&	0.17	&	0.21	&	0.31	& $	[1.09E4, 8.91E7, 3.37E8]	$ \\	 MeshGraphNets~\cite{pfaff2020learning} & 	0.034	&	0.065	&	0.071	&	0.064	& $	[1.69E4, 1.40E5, 8.45E5]	$ \\
    \hline
    \end{tabular}
    \caption{Quantitative results: Ortiz, 326-frame jazz dancing.}
    \label{tab:ortiz_2}
\end{table*}

\begin{table*}[!htbp]
    \centering
    \begin{tabular}{l|c|c|c|c|c|c}
    \hline
     character & \makecell{\# vertices \\ (tet mesh)} & \makecell{$t_{GT}$ \\ s/frame \\ 1 step / frame} & \makecell{$t_{BL}$ \\ s/frame \\ 1 step / frame } & \makecell{$t_{BL}$ \\ s/frame \\ 50 steps / frame } & \makecell{$t_{BL}$ \\ s/frame \\ 100 steps / frame } & \makecell{$t_{ours}$ \\ s/frame \\ 1 step / frame }  \\
     \hline
     Big vegas & 1468 & 0.58 & 0.056 & 2.57027 & 6.20967 & \textbf{0.017}\\
     Kaya & 1417 & 0.52	& 0.052 & 2.42985 & 5.72762 & \textbf{0.015}\\
     Michelle & 1105 & 0.33 & 0.032 & 1.52916 & 3.64744 & \textbf{0.013}\\
     Mousey & 2303 & 0.83 & 0.084 & 3.90579 & 9.5897 &  \textbf{0.018}\\
     Ortiz & 1258 & 0.51 & 0.049 & 2.2496 & 5.16806 & \textbf{0.015} \\
     \hline
    \end{tabular}
    \caption{The running time of the ground truth, the baseline, and our method.}
    \label{tab:time_result_detail}
\end{table*}

\begin{table*}[!htbp]
    \centering
    \begin{tabular}{c|l|c|c|c|c}
    \hline
    Motion & Dataset & Single Frame & Rollout-24 & Rollout-48 & Rollout-All \\
    \hline
    \multirow{2}{*}{\makecell{	Hip hop dancing 1	\\	283 frames	}}	&Sphere Dataset&	\textbf{0.0098}	 & 	0.053	 & 	\textbf{0.063}	 & 	\textbf{0.0591}\\ 
    & Ortiz Dataset  & 	0.0111	 & 	\textbf{0.0512}	 & 	0.067	 & 	0.0969	\\\hline
    \multirow{2}{*}{\makecell{	Hip hop dancing 2	\\	366 frames	}}	&Sphere Dataset &	\textbf{0.0093}	 & 	0.0664	 & 	0.0744	 & 	\textbf{0.0727}	\\ 
    & Ortiz Dataset  & 	0.0101	 & 	\textbf{0.0564}	 & 	\textbf{0.0607}	 & 	0.1241	\\\hline
    \multirow{2}{*}{\makecell{	Samba dancing 1	\\	594 frames	}}	&Sphere Dataset &	\textbf{0.0062}	 & 	0.0495	 & 	\textbf{0.0571}	 & 	\textbf{0.0654}	\\ 
    & Ortiz Dataset  & 	0.0062	 & 	\textbf{0.0481}	 & 	0.061	 & 	0.143	\\ \hline
    \multirow{2}{*}{\makecell{	Samba dancing 2	\\	493 frames	}}	&Sphere Dataset &	\textbf{0.0058}	 & 	0.0521	 & 	0.0496	 & 	\textbf{0.0635}	\\ 
    & Ortiz Dataset  & 	0.0064	 & 	\textbf{0.0367}	 & 	\textbf{0.0423}	 & 	0.1331	\\ \hline
    \multirow{2}{*}{\makecell{	Samba dancing 3	\\	399 frames	}}	&Sphere Dataset &	\textbf{0.0065}	 & 	0.0615	 & 	\textbf{0.0537}	 & 	\textbf{0.0702}	\\ 
    & Ortiz Dataset  & 	0.0065	 & 	\textbf{0.0383}	 & 	0.0616	 & 	0.1282	\\
    \hline
    \end{tabular}
    \caption{Quantitative results: the network trained on different datasets and tested on the character Big vegas.}
    \label{tab:dataset_big_vegas}
\end{table*}

\begin{table*}[!htbp]
    \centering
    \begin{tabular}{c|l|c|c|c|c}
    \hline
    Motion & Dataset & Single Frame & Rollout-24 & Rollout-48 & Rollout-All \\
    \hline
    \multirow{2}{*}{\makecell{	Dancing running man	\\	650 frames	}}	&Sphere Dataset &	\textbf{0.0067}	 & 	\textbf{0.0544}	 & 	\textbf{0.0578}	 & 	\textbf{0.0411}	\\ 
    & Ortiz Dataset  & 	0.0113	 & 	0.075	 & 	0.0839	 & 	0.1176	\\ \hline
\multirow{2}{*}{\makecell{	Zombie scream	\\	167 frames	}}	&Sphere Dataset &	\textbf{0.0075}	 & 	0.0834	 & 	0.0666	 & 	\textbf{0.0537}	\\ 
& Ortiz Dataset  & 	0.0126	 & 	\textbf{0.0749}	 & 	\textbf{0.0645}	 & 	0.076	\\
    \hline
    \end{tabular}
    \caption{Quantitative results: the network trained on different datasets and tested on the character Kaya.}
    \label{tab:dataset_kaya}
\end{table*}

\begin{table*}[!htbp]
    \centering
    \begin{tabular}{c|l|c|c|c|c}
    \hline
    Motion & Dataset & Single Frame & Rollout-24 & Rollout-48 & Rollout-All \\
    \hline
    \multirow{2}{*}{\makecell{	Gangnam style	\\	371 frames	}}	&Sphere Dataset &	\textbf{0.0041}	 & 	\textbf{0.0329}	 & 	0.0332	 & 	\textbf{0.0401}	\\ 
    & Ortiz Dataset  & 	0.0059	 & 	0.0329	 & 	\textbf{0.0319}	 & 	0.0969	\\ \hline
    \multirow{2}{*}{\makecell{	Swing dancing 1	\\	627 frames	}}	&Sphere Dataset &	\textbf{0.0056}	 & 	\textbf{0.025}	 & 	\textbf{0.0236}	 & 	\textbf{0.0471}	\\ 
    & Ortiz Dataset  & 	0.0079	 & 	0.0324	 & 	0.035	 & 	0.1213	\\ \hline
    \multirow{2}{*}{\makecell{	Swing dancing 2	\\	699 frames	}}	&Sphere Dataset &	\textbf{0.0056}	 & 	\textbf{0.0491}	 & 	\textbf{0.0373}	 & 	\textbf{0.0548}	\\ 
    & Ortiz Dataset  & 	0.0067	 & 	0.058	 & 	0.04	 & 	0.1264	\\
    \hline
    \end{tabular}
    \caption{Quantitative results: the network trained on different datasets and tested on the character Michelle.}
    \label{tab:dataset_michelle}
\end{table*}

\begin{table*}[!htbp]
    \centering
    \begin{tabular}{c|l|c|c|c|c}
    \hline
    Motion & Dataset & Single Frame & Rollout-24 & Rollout-48 & Rollout-All \\
    \hline
    \multirow{2}{*}{\makecell{	Dancing	\\	158 frames	}}	&Sphere Dataset &	\textbf{0.008}	 & 	\textbf{0.0771}	 & 	\textbf{0.1003}	 & 	\textbf{0.0858}	\\ 
    & Ortiz Dataset  & 	0.0122	 & 	0.0871	 & 	0.1056	 & 	0.122	\\ \hline
    \multirow{2}{*}{\makecell{	Shuffling	\\	225 frames	}}	&Sphere Dataset &	\textbf{0.0066}	 & 	\textbf{0.0666}	 & 	0.1115	 & 	\textbf{0.09}	\\ 
    & Ortiz Dataset  & 	0.0115	 & 	0.0936	 & 	\textbf{0.1072}	 & 	0.1303	\\ \hline
    \multirow{2}{*}{\makecell{	Swing dancing	\\	627 frames	}}	&Sphere Dataset &	\textbf{0.009}	 & 	\textbf{0.0871}	 & 	\textbf{0.1001}	 & 	\textbf{0.1042}	\\
    & Ortiz Dataset  & 	0.0144	 & 	0.1236	 & 	0.1113	 & 	0.1594	\\ 
    \hline
    \end{tabular}
    \caption{Quantitative results: the network trained on different datasets and tested on the character Mousey.}
    \label{tab:dataset_mousey}
\end{table*}

\begin{table*}[h]
    \centering
    \begin{tabular}{c|l|c|c|c|c}
    \hline
    Motion & Dataset & Single Frame & Rollout-24 & Rollout-48 & Rollout-All \\
    \hline
    \multirow{2}{*}{\makecell{	cross jumps rotation	\\	122 frames	}}	& Sphere Dataset &	0.0057	 & 	0.0819	 & 	0.0765	 & 	0.0726	\\ 
    & Ortiz Dataset  & 	\textbf{0.0053}	 & 	\textbf{0.0719}	 & 	\textbf{0.0532}	 & 	\textbf{0.0702}	\\ 
    \hline
    \multirow{2}{*}{\makecell{	jazz dancing	\\	326 frames	}}	& Sphere Dataset &	\textbf{0.0039}	 & 	\textbf{0.0391}	 & 	\textbf{0.0316}	 & 	\textbf{0.0363}	\\
     & Ortiz Dataset  & 	0.0051	 & 	0.0365	 & 	0.0338	 & 	0.0946	\\ 
    \hline
    \end{tabular}
    \caption{Quantitative results: the network trained on different datasets and tested on the character Ortiz.}
    \label{tab:dataset_ortiz}
\end{table*}

\begin{table*}[h]
    \centering
    \begin{tabular}{c|l|c|c|c|c}
    \hline
    Test Dataset & Patch size & Single Frame & Rollout-24 & Rollout-48 & Rollout-All \\
    \hline
    \multirow{3}{*}{Big vegas}	& 1-ring &	0.0075	 & 0.057	 & 0.060	 & 	0.066 \\ &    2-ring  & 	\textbf{0.0068}	 & 0.060	 & 	0.066	 & 	0.077\\ &   3-ring & 0.0075	 & 	\textbf{0.043} & \textbf{0.0418}	 & \textbf{0.059}	\\    \hline
    \multirow{3}{*}{Kaya}	& 1-ring &	0.0071	 & \textbf{0.069}	 & 	0.062 & \textbf{0.047}	\\ &    2-ring  & \textbf{0.0060} & 0.078	 & 	\textbf{0.060}	 & 	0.072	\\ &   3-ring & 0.0064	 & 0.099	 & 	0.095	 & 0.092	\\    \hline
    \multirow{3}{*}{Michelle}	& 1-ring &	0.0051	 & 0.036	 & 		\textbf{0.031} & 0.047	\\ &    2-ring  & 	\textbf{0.0045}	 & 	\textbf{0.033} & 	0.033	 & 	\textbf{0.047}	\\ &   3-ring & 0.0047	 & 0.043	 & 	0.042 & 0.059	\\    \hline
    \multirow{3}{*}{Mousey}	& 1-ring &	0.0079	 & 	\textbf{0.077} & \textbf{0.10}	 & \textbf{0.093}	\\ &    2-ring  & \textbf{0.0069} & 	0.11 & 	0.19	 & 	0.14	\\ &   3-ring & 	0.0075	 & 		0.11 & 	0.21 & 0.16	\\    \hline
    \multirow{3}{*}{Ortiz}	& 1-ring &		0.0048	 & 	\textbf{0.061}	 & 	\textbf{0.054} & \textbf{0.054}	\\ &    2-ring  & 	\textbf{0.0042}	 & 	0.069	 & 	0.089	 & 0.070		\\ &   3-ring & 0.0051	 & 0.094	 & 0.097	 & 	0.087\\    \hline
    \end{tabular}
    \caption{Quantitative results: the network trained on local patches of different patch sizes.}
    \label{tab:ring_size_result}
\end{table*}


\end{document}